\documentclass[final,5p,times,twocolumn]{elsarticle}

\usepackage{graphicx} 

\usepackage{subcaption}
\usepackage{amssymb}
\usepackage{amsmath}

\usepackage{float}
\usepackage{siunitx}
\usepackage[utf8]{inputenc}   
\usepackage{hyperref}
\usepackage{cleveref}
\usepackage{comment}
\usepackage{booktabs}  
\usepackage{tabularx}  
\usepackage[inline]{enumitem} 
\usepackage{xcolor}
\usepackage{interval}
\usepackage[
nolist,nohyperlinks
]{acronym}
\usepackage{xurl}

\journal{Robotics and Autonomous Systems}



\begin{document}


\acrodefplural{OS}[OS's]{Operating Systems}
\acrodefplural{MAV}[MAVs]{Micro Aerial Vehicles}

\begin{acronym}[Dueling DQN]
    
\acro{A2C}[A2C]{Advantage Actor-Critic}
\acro{A3C}[A3C]{Asynchronous Advantage Actor-Critic}
\acro{BCD}[BCD]{Boustrophedon Cellular Decomposition}
\acro{CG}[CG]{Constraint Graph}
\acro{CNN}[CNN]{Convolutional Neural Network}
\acro{CSLAM}[C-SLAM]{Collaborative-SLAM}
\acro{CTCE}[CTCE]{Centralized Training Centralized Execution}
\acro{CTDE}[CTDE]{Centralized Training Decentralized Execution}
\acro{D3QN}[D3QN]{Dueling Double Deep Q-Networks}
\acro{Dec-POMDP}[Dec-POMDP]{Decentralized Partially Observable Markov Decision Process}
\acro{DCOP}[DCOP]{Distributed Constrained Optimization Problem}
\acro{DDQN}[DDQN]{Double Deep Q-Network}
\acro{DOGM}[DOGM]{Dynamic OGM}
\acro{DRL}[DRL]{Deep Reinforcement Learning}
\acro{DQL}[DQL]{Deep Q-Learning}
\acro{DTDE}[DTDE]{Decentralized Training Decentralized Execution}
\acro{Dueling DQN}[Dueling DQN]{Dueling Deep Q-Networks}
\acro{EWMA}[EWMA]{Exponential Weighted Moving Average}
\acro{FoV}[FoV]{Field of View}
\acro{GAE}[GAE]{Generalized Advantage Estimation}
\acro{GNN}[GNN]{Graph Neural Network}
\acro{GNSS}[GNSS]{Global Navigation Satellite System}
\acro{GRU}[GRU]{Gated Recurrent Unit}
\acro{HPC}[HPC]{High-Performance Computing}
\acro{HRL}[HRL]{Hierarchical Reinforcement Learning}
\acro{INS}[INS]{Integrated Navigation System}
\acro{IPPO}[IPPO]{Independent Proximal Policy Optimization}
\acro{IRL}[IRL]{Inverse Reinforcement Learning}
\acro{LiDAR}[LiDAR]{Light Detection and Ranging}
\acro{LSTM}[LSTM]{Long Short-Term Memory}
\acro{MA}[MA]{Multi Agent}
\acro{MACPP}[MACPP]{Multi-Agent Coverage Path Planning}
\acro{MADDPG}[MADDPG]{Multi-Agent Deep Deterministic Policy Gradient}
\acro{MAPPO}[MAPPO]{Multi-Agent PPO with a centralized critic}
\acro{MARL}[MARL]{Multi-Agent Reinforcement Learning}
\acro{MAV}[MAV]{Micro Aerial Vehicles}
\acro{MCTS}[MCTS]{Monte Carlo Tree Search}
\acro{MDP}[MDP]{Markov Decision Process}
\acro{ND-POMDP}[ND-POMDP]{Network Distributed POMDP}
\acro{NE}[NE]{Nash Equilibrium}
\acro{NN}[NN]{Neural Network}
\acro{OGM}[OGM]{Occupancy Grid Map}
\acro{POMDP}[POMDP]{Partially Observable MDP}
\acro{POSG}[POSG]{Partially Observable Stochastic Game}
\acro{PPO}[PPO]{Proximal Policy Optimization}
\acro{QL}[QL]{Q-Learning}
\acro{QMIX}[QMIX]{Q-value Mixing}
\acro{RL}[RL]{Reinforcement Learning}
\acro{RNN}[RNN]{Recurrent Neural Network}
\acro{RRT}[RRT]{Rapidly-exploring Random Tree}
\acro{SLAM}[SLAM]{Simultaneous Localization And Mapping}
\acro{SLURM}[SLURM]{Simple Linux Utility for Resource Management}
\acro{UAV}[UAV]{Unmaned Aerial Vehicle}


\end{acronym}

\begin{frontmatter}



\title{IMAGINE: Intelligent Multi-Agent Godot-based Indoor Networked Exploration}

\author[inesc-inov]{Tiago Leite\corref{cor}} 
\ead{tiagomleite@tecnico.ulisboa.pt}

\author[inesc-inov,inesc-id,isr]{Maria Inês Conceição} 
\ead{ines.conceicao@tecnico.ulisboa.pt}

\author[inesc-inov]{António Grilo}
\ead{antonio.grilo@inov.pt}

\cortext[cor]{Corresponding author.}

\affiliation[inesc-inov]{organization={INESC INOV, Instituto Superior Técnico, Universidade de Lisboa},
             city={Lisbon},
             postcode={1000-029},
             country={Portugal}}

\affiliation[inesc-id]{organization={
INESC ID–Instituto de Engenharia de Sistemas e Computadores: Investigação e Desenvolvimento, Instituto Superior Técnico},
             city={Lisbon},
             postcode={1000-029},
             country={Portugal}}

\affiliation[isr]{organization={Institute for Systems and Robotics, Instituto Superior Técnico, Universidade de Lisboa},
             city={Lisbon},
             postcode={1049-001},
             country={Portugal}}







\begin{abstract}

The exploration of unknown, Global Navigation Satellite System (GNSS) denied environments by an autonomous communication-aware and collaborative group of Unmanned Aerial Vehicles (UAVs) presents significant challenges in coordination, perception, and decentralized decision-making.
This paper implements Multi-Agent Reinforcement Learning (MARL) to address these challenges in a  2D indoor environment, using high-fidelity game-engine simulations (Godot) and continuous action spaces.
Policy training aims to achieve emergent collaborative behaviours and decision-making under uncertainty using Network-Distributed Partially Observable Markov Decision Processes (ND-POMDPs).
Each UAV is equipped with a Light Detection and Ranging (LiDAR) sensor and can share data (sensor measurements and a local occupancy map) with neighbouring agents.
Inter-agent communication constraints include limited range, bandwidth and latency.
Extensive ablation studies evaluated MARL training paradigms, reward function, communication system, neural network (NN) architecture, memory mechanisms, and POMDP formulations.
This work jointly addresses several key limitations in prior research, namely reliance on discrete actions, single-agent or centralized formulations, assumptions of \textit{a priori} knowledge and permanent connectivity, inability to handle dynamic obstacles, short planning horizons and architectural complexity in Recurrent NNs/Transformers.
Results show that the scalable training paradigm, combined with a simplified architecture, enables rapid
autonomous exploration of an indoor area.
The implementation of Curriculum-Learning (five increasingly complex levels) also enabled faster, more robust training.
Area coverage surpassed 95\% in three
levels, and the implementation of Convolutional NNs (CNNs) increased
system performance by 20\%.
This combination of high-fidelity simulation, MARL formulation, and computational efficiency establishes a strong foundation for deploying learned cooperative strategies in physical robotic systems.

\end{abstract}


\end{frontmatter}



\section{Introduction}
\label{sec:intro}

Initially developed for military purposes, \acp{UAV} have now become increasingly popular in a wide range of applications, including commercial use, security~\cite{uav-monitor}, surveillance~\cite{uav_surveilance}, cinema~\cite{uav_cinema}, agriculture~\cite{uav_agriculture}, disaster management~\cite{uav-search-rescue}, and personal use~\cite{more-drones}.


The deployment of \ac{UAV} teams has gained attention recently, with collaboration among \acp{UAV} being a highly desirable, albeit challenging, capability.
This is inspired by natural systems where cooperation and decentralization are essential for solving complex tasks.


This work is positioned within the context of utilizing swarms of robots, particularly \acp{UAV} and potentially \acp{MAV}, for indoor exploration.


Specifically, the problem formulation is using multiple decentralized agents to explore an \textit{a priori} unknown indoor 2D environment,
whilst being able to communicate amongst themselves, given realistic network constraints.

The problem of cooperative multi-agent exploration is challenging due to environmental uncertainty and communication constraints. A comparative summary of the most relevant related works is provided in \Cref{tab:related_work}.
Methodological paradigms offer distinct trade-offs.
\textbf{Frontier-based} methods direct agents to unknown region boundaries~\cite{frontier-based} and use map-sharing for coordination~\cite{multiagent-octomap, ivfm, sgraphs-cslam}, but can be myopic.
\textbf{Planning-based} approaches use cellular decomposition~\cite{planning_bcd, planning_voronoi, planning_Morse} for systematic coverage, requiring a global model.
\textbf{Sampling-based} algorithms like \ac{RRT}~\cite{sampling_rrtstar, sampling_cstar} explore complex spaces but share a high communication burden.

\textbf{RL-based} methods learn exploration policies, either in a two-stage manner with classical planners~\cite{2stage_rl_bcd} or end-to-end~\cite{SelfLearningDRL, learning_cp_unkown_env_drl}.
However, \ac{MARL} introduces coordination challenges, often addressed with computationally expensive techniques~\cite{marl-joint-action-variable-elimination, gnn-marl1, gnn-marl3}.
A significant gap remains for communication-efficient, multi-agent strategies under full uncertainty.

\begin{table*}[!htbp] 
    \centering

    \caption{Comparison of key related works in multi-agent exploration.}
    \label{tab:related_work}
    \renewcommand\tabularxcolumn[1]{m{#1}}

\newcolumntype{Y}{>{\centering\arraybackslash}X}

\begin{tabularx}{\textwidth}{cp{2.2cm}YYc}
\toprule
Ref & Name & Problem & Solution & MA \\
\midrule
\cite{SelfLearningDRL} & Self-Learn DRL & Unkown map exploration by finding the best sensing action. & DQL with and without RNNs. A CNN outputs the best sensing action from a local map and Frontier Rescue when the local map is completely known. & no \\ \hline
\cite{ivfm} & i-VFM & Reduce Multi Agent repetitve exploration & State representation (map), named i-VFM, that integrates the number of previous agent visits. & yes \\ \hline
\cite{multiagent-octomap} & MA Octomap & Autonomous exploration of unknown subterranean environments. & Navigation using a shared map with Graph-based and Frontier-based planning. & yes \\ \hline
\cite{sgraphs-cslam} & Multi S-Graphs & Collaboratively and efficiently build a map with multiple agents. & Semantic Graphs. & yes \\
\bottomrule
\end{tabularx}

    {\raggedright \textit{Abbreviations: MA - Multi-Agents.} \par}
\end{table*}


This work integrates in a single framework several aspects already present in the state-of-the-art, which can be considered a contribution:
\begin{enumerate}[label=(\arabic*)]
\item Adopting the separation of concerns into mapping and decision-making, as demonstrated in~\cite{multiagent-octomap}.
\item Leveraging coordination mechanisms at the map level, inspired by~\cite{sgraphs-cslam}, and with sensor information as in~\cite{gnn-marl1,gnn-marl3}, thereby avoiding reliance on computationally expensive techniques.
\item Building upon the end-to-end \ac{DRL} paradigm and extend it to the multi-agent domain, employing a continuous action space as in~\cite{learning_cp_unkown_env_drl}, unlike the discrete space in~\cite{SelfLearningDRL}.
\item Integrating strategies from~\cite{SelfLearningDRL, learning_cp_unkown_env_drl} in reward formulation.
\end{enumerate}

Nevertheless, it also brings new contributions:

\begin{enumerate}[label=(\arabic*)]
\item It adopts a \ac{MARL} paradigm for UAV swarm coordination in exploration missions, addressing this current gap in the literature.
\item It assesses the impact of communications in mission coordination by implementing a more realistic communication model, including data rate and range limitations.  
\item Instead of defaulting to \acp{RNN} for partial observability, this work successfully employs a Belief-\ac{MDP} formulation, providing a more structured and effective approach to state estimation.
\item It strives to keep physical realism (while also maintaining computational efficiency) by simulating the environment using the open-source game engine Godot~\cite{godotengine}, which, to the best of the authors' knowledge, is the first time this engine is applied in this field. Godot provides a lightweight, open-source platform for high-fidelity, game-engine-quality simulation, complete with physically accurate collisions for both static and dynamic objects.
\item Utilizing industry-grade tools (Ray/RLlib, WandB), with validation on an \ac{HPC} cluster, ensuring reproducibility and scalability.
\end{enumerate}


The remainder of this document is structured as follows.\newline
\Cref{sec:backg} covers the theoretical background on \ac{ND-POMDP}.
\Cref{sec:imple} describes the proposed method and its practical implementation.
\Cref{sec:resul} presents the experimental results.
Finally, \Cref{sec:concl} presents the main conclusions and directions for future work.


\section{Network Distributed-Partially Observable Markov Decision Process}
\label{sec:backg}

The standard \ac{MARL} Mathematical framework is the \ac{Dec-POMDP}, defined by the tuple $(\mathcal{N},\mathbb{S},\mathbb{A},\mathbb{O},O,T,R,\gamma)$.
The \ac{Dec-POMDP} can be further extended to an \ac{ND-POMDP} by modeling communication constraints between agents using a graph $\mathbb{G}$.
Therefore, an \ac{ND-POMDP} is defined by the tuple $(\mathcal{N}, \mathbb{S}, \mathbb{A}, T, R, \mathbb{O}, O, \mathbb{G}, \gamma)$:
\begin{itemize}
    \item $\mathcal{N}$ is the set of agents, indexed by $i \in \mathcal{N}$ where the total number of agents is $n = |\mathcal{N}|$.
    \item $\mathbb{S}$ is the state space, the set of possible states.
    \item $\mathbb{A}$, where $\mathbb{A}_i$ is the set of actions available to agent $i$ which can be different for each agent, and $\mathbb{A} = \mathbb{A}_1 \times \mathbb{A}_2 \times \dots \times \mathbb{A}_{|\mathcal{N}|}$ is the joint action space.
    At each time step $t$, each agent $i$ takes an action\footnote{
        This work adopts the convention that time step subscripts may be omitted when unambiguous.
        For instance, an action at time $t$ for agent $i \in \mathcal{N}$ can be denoted explicitly as $a^t_i$ or implicitly as $a_i$.
        Similarly, relative temporal notation is used, where $s$ denotes the current state, $a$ the selected joint action, and $s'$ the successor state.
        This allows the reward function to be expressed compactly as $R(s, a, s')$ instead of $R(s^t, a^t, s^{t+1})$.
    } $a_{i}^t$, leading to one joint action $a = ( a_1,\dots,a_{|\mathcal{N}|} )$ at each time step~\cite{intro_decpomp}.
    \item $T: \mathbb{S} \times \mathbb{S} \times \mathbb{A} \rightarrow \mathbb{R}$ is the state transition probability function $T(s', s, a)=P(s' | s, a)$, describing the probability of transitioning to state $s'$ given current state $s$ and joint action $a = ( a_1, a_2, \dots, a_n )$.
    \item $R: \mathbb{S} \times \mathbb{A} \rightarrow \mathbb{R}$ is the shared reward function $R(s, a, s')$, providing the same scalar reward to all agents based on the current state $s$, joint action $a$ and the environment transitioning to state $s'$.
    \item $\mathbb{O}$, where $\mathbb{O}_i$ is the set of observations available to agent $i$ which can be different for each agent, and $\mathbb{O} = \mathbb{O}_1 \times \mathbb{O}_2 \times \dots \times \mathbb{O}_{|\mathcal{N}|}$ is the joint observation space.
    \item $O: \mathbb{O} \times \mathbb{S} \times \mathbb{A} \rightarrow \mathbb{R}$ is the joint observation probability function $O(o, s', a)=P(o | s', a)$, describing the probability of observing $o = ( o_1, o_2, \dots, o_{|\mathcal{N}|} )$ given the next state $s'$ and joint action $a$.
    \item $\mathbb{G}$: The set of all possible communication graphs, where each graph $g=(V, E)$ represents a communication topology at a given time.
    The vertex set $V$ corresponds to the agents $\mathcal{N}$, and the edge set $E$ represents communication links.
    At each time step $t$, the current communication graph\footnote{The possible communication configurations grow exponentially with the number of agents $|\mathbb{G}|=2^{n \choose 2}=2^{n(n-1)/2}$ for $n = |\mathcal{N}|$ agents.} $g_t \in \mathbb{G}$ contains an edge $(i, j)$ if and only if agents $i$ and $j$ can communicate at time $t$.
    \item $\gamma \in \interval{0}{1}$ is the discount factor, determining the importance of future rewards.
\end{itemize}

Two primary approaches for solving these complex multi-agent problems are:
\begin{enumerate*}[label=(\roman*)]
    \item Formulating them as \acp{DCOP}, or
    \item Leveraging the structure of the communication graph using \acp{GNN},
    which allow agents to efficiently aggregate and process information from their neighbors~\cite{gnn-marl1,gnn-marl2,gnn-marl3}.
\end{enumerate*}

The approach proposed in this paper will not follow either of these options.
It will simplify the problem by allowing neighboring agents to share observations.
In this way, it avoids the computational complexity, whilst still benefitting from communication and coordination between agents.


\section{Proposed Method}
\label{sec:imple}

The implemented system architecture is comprised of two core components: the Godot-based environment simulation and the Python-based learning algorithm using Ray/RLlib, connected via the Godot RL Agents library, shown in~\Cref{fig:python_godot_process_diagram_v3}.
The complete implementation is available in the GitHub repository~\cite{githubIMAGINE}.


\subsection{Architecture}
\begin{figure}[t]
    \centering
    \includegraphics[width=0.4\textwidth]{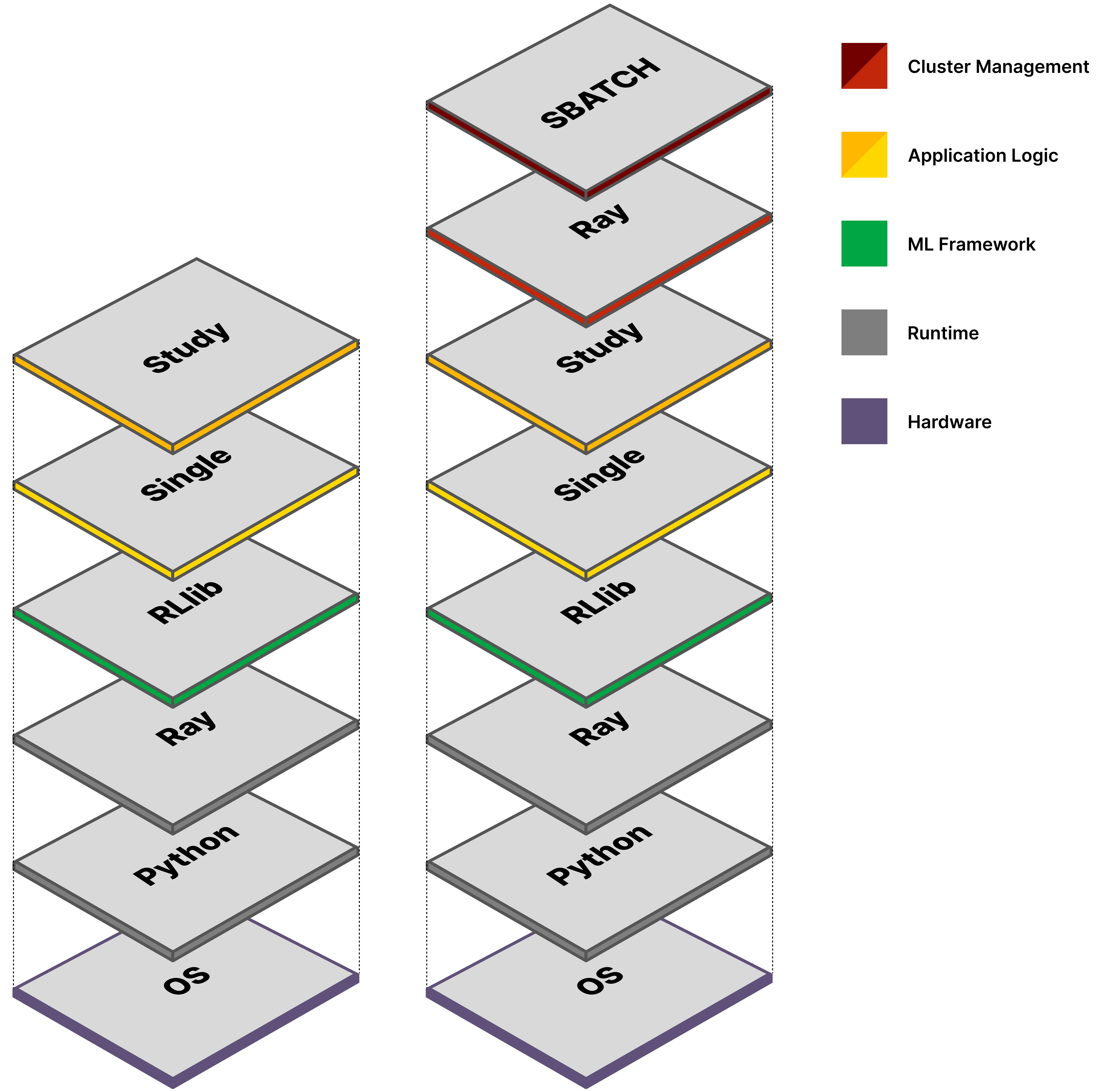}
    \caption{Architecture Stack.}
    \label{fig:stack_diagram}
\end{figure}

The Python-side implementation provides a foundation for large-scale experimentation:

\begin{itemize}
\item Single Trial: Configurable via a YAML file containing all hyperparameters (e.g., training batch size, number of agents, paradigm, etc.).

\item Multi-Trial: Leverage Ray's distributed framework for parallel execution of grid searches, specific configurations, and hyperparameter tuning.

\item Study: Preconfigured Multi-Trials.

\item Hardware Abstraction: Ensures portability across \ac{HPC} clusters (Deucalion and Cirrus) through environment detection and UV package management.

\item Experiment Tracking: Weights \& Biases~\cite{wandb} Integration for real-time monitoring of $\sim$200 metrics per trial.

\begin{figure}[!htbp]
    \centering
    \includegraphics[width=0.45\textwidth]{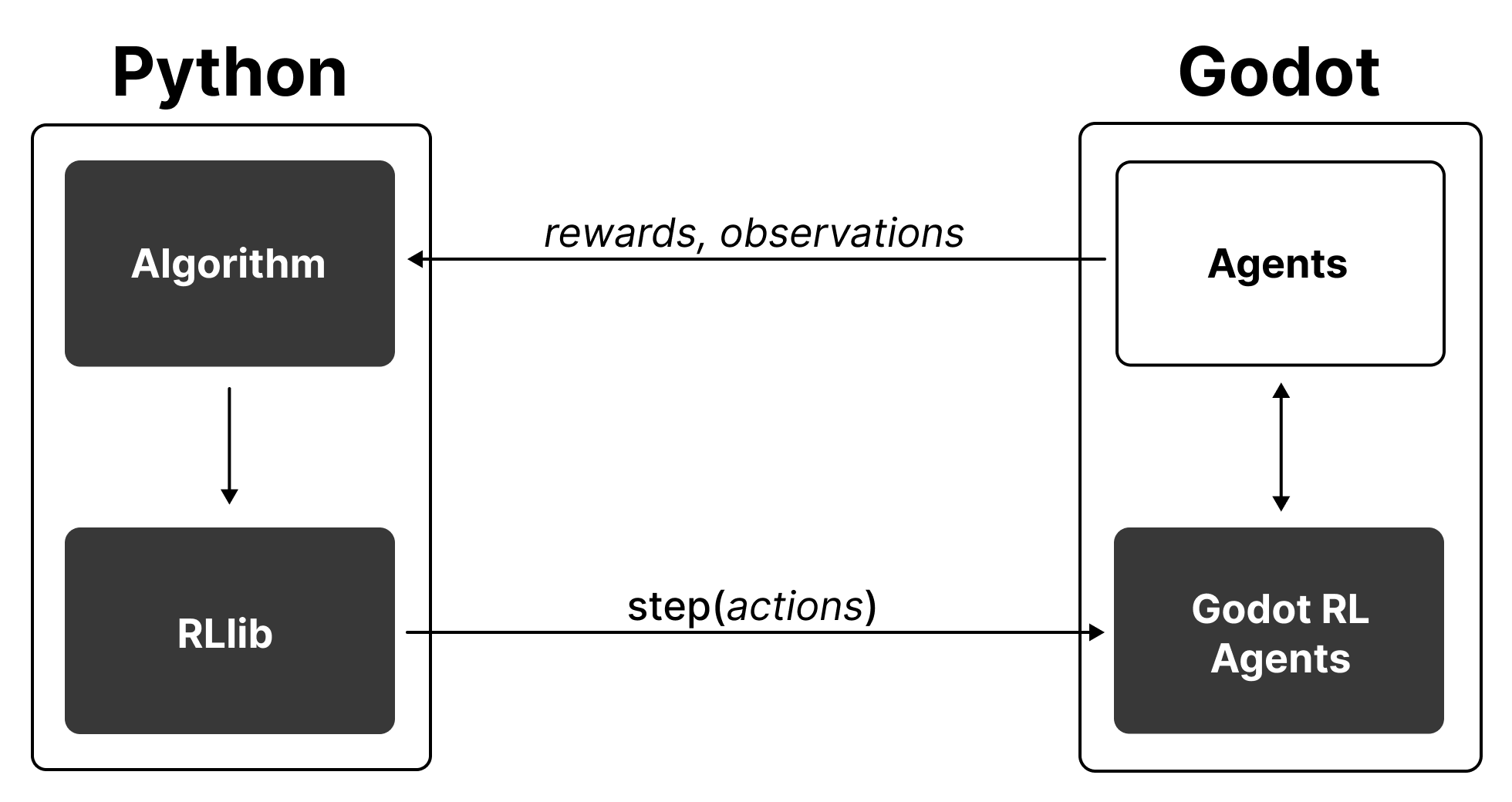}
    \caption{Godot-Python RL Process.}
    \label{fig:python_godot_process_diagram_v3}
\end{figure}

\end{itemize}
This complete architecture stack, illustrating both local and \ac{HPC} execution paths, is summarized in~\Cref{fig:stack_diagram}.

\begin{figure}[htbp]
    \begin{subfigure}[t]{0.49\linewidth}
        \centering
        \includegraphics[width=\linewidth]{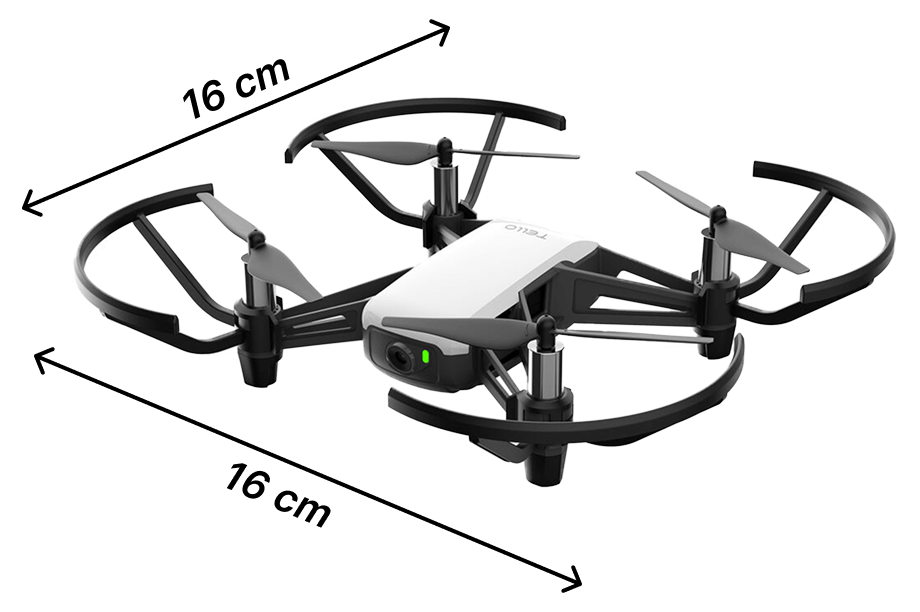}
        \small
        \caption{Real world DGI Tello UAV.}
        \label{fig:drone_scale_diagram}
    \end{subfigure}
    \begin{subfigure}[t]{0.49\linewidth}
        \centering
        \includegraphics[width=\linewidth]{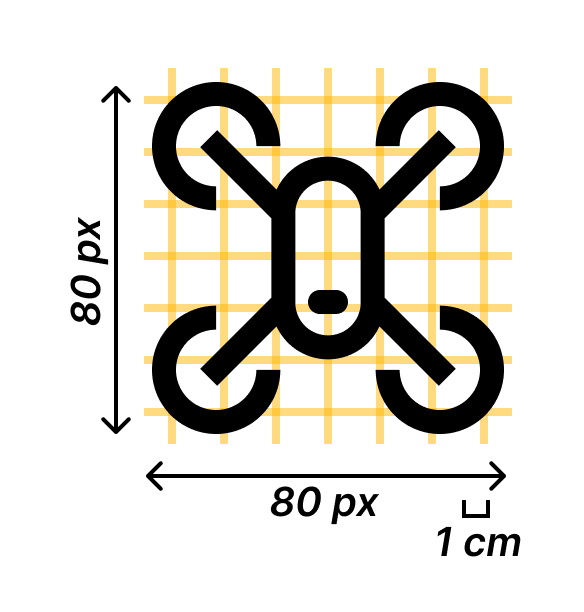}
        \small
        \caption{Simulated UAV.}
        \label{fig:drone_scale_diagram_godot}
    \end{subfigure}
    \caption{Real world UAV and its simulated counterpart.}
\end{figure}

\textbf{Environment Simulation}
is built using the Godot game engine~\cite{godotengine}, which provides a physically-grounded simulation with realistic constraints.
With no loss of generality, the simulated \ac{UAV}, controlled by the previously mentioned Python framework, is modeled after the DJI Tello.
A comparison of the real-world \ac{UAV} and its simulated counterpart is shown in~\Cref{fig:drone_scale_diagram,fig:drone_scale_diagram_godot}.

With a simulation scale of $\frac{\SI{16}{cm}}{\SI{80}{px}}=\SI{0.2}{cm/px}$,
the \ac{UAV} Kinematics are also modeled after DJI Tello specifications with tuned velocities (\SI{0.8}{m/s} linear, \SI{3}{rad/s} angular).

\begin{figure}[!]
    \centering
    \includegraphics[width=0.50\textwidth]{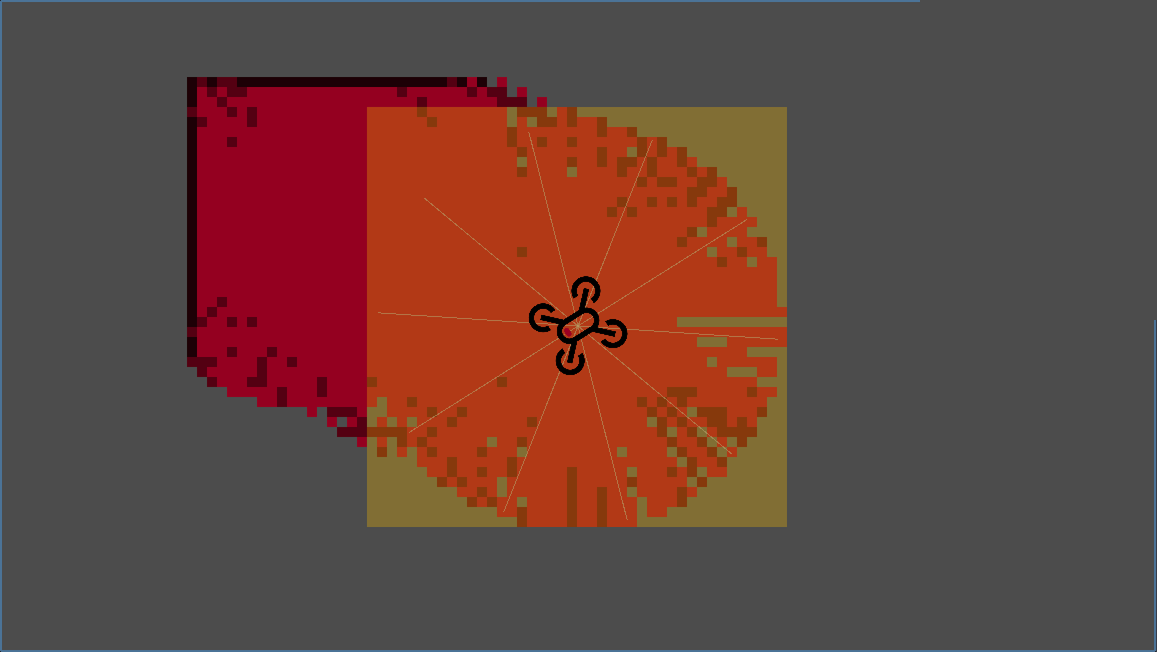}
    \caption{UAV sensing and mapping:
        (1) Simulated \ac{LiDAR} rays (grey lines),
        (2) The Local Map of the agent (red area), and
        (3) The Egocentric Map provided to the policy network (yellow area).}
    \label{fig:drone_lidar_obs_map}
\end{figure}

It is assumed that the \ac{UAV} model is equipped with a single sensor: a \ac{LiDAR}.
The \ac{LiDAR} is simulated using RayCast2D nodes with limited range and field of view.
This information is used to update its \textit{Local Map}, an Occupancy Grid Map, that utilizes Bayesian log-odds updates. 
The policy receives as input a fixed-size \textit{Egocentric Map}, which is a portion of the \textit{Local Map} centered on the agent.
The integrated sensing and mapping pipeline is illustrated in~\Cref{fig:drone_lidar_obs_map}, showing \ac{LiDAR} rays, color-modulated Local Map, and the Egocentric Map provided to policy networks.

\begin{figure}[!]
    \centering
    \includegraphics[width=0.50\textwidth]{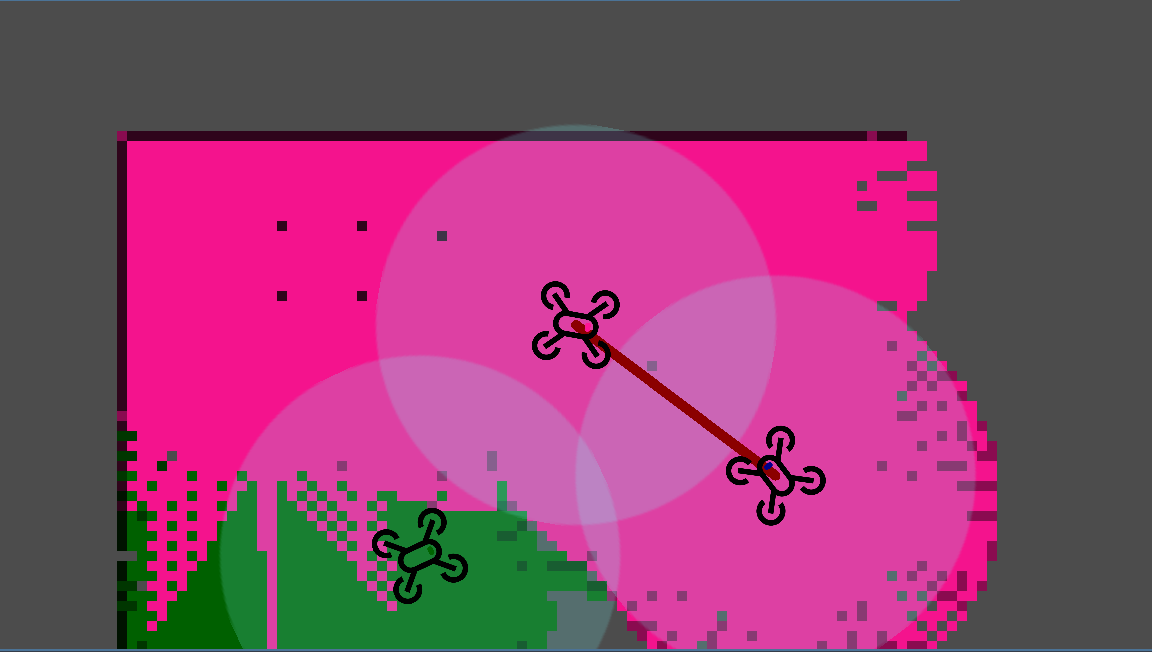}
    \caption{UAV network: communication ranges (blue circles), active links (red lines), and agent Local Maps (pink/green).}
    \label{fig:drone_network_conn_range}
\end{figure}

The \acp{UAV} are capable of communicating using limited-range networking (depicted in~\Cref{fig:drone_network_conn_range}) with simulated transmission delays based on message size and bandwidth.
They use a simple single-hop protocol, continuously share \ac{LiDAR} data with agents in range, and share the entire \textit{Local Map} upon (re)-establishing connection.

\begin{figure}[!htbp]
    \centering
    \includegraphics[width=0.45\textwidth]{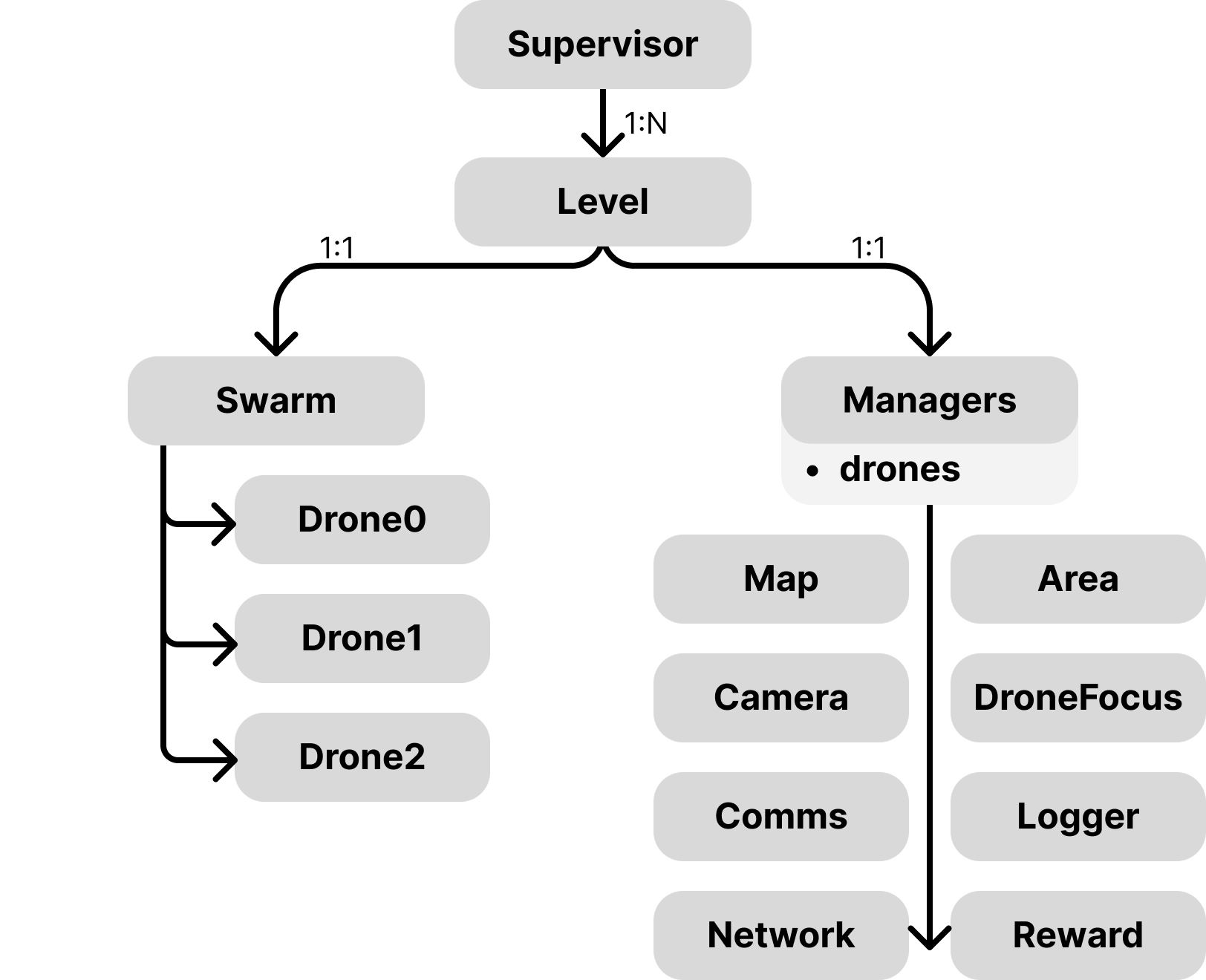}
    \caption{Godot Architecture.}
    \label{fig:godot_node_architecture}
\end{figure}

The modular Godot architecture, depicted in~\Cref{fig:godot_node_architecture}, separates concerns through specialized managers for swarm control, communication, reward calculation, and logging.

\pagebreak

\textbf{MARL Paradigms Implementation}.
\ac{MARL} paradigms are distinguished by their approach across two phases: training and execution.
During training, the objective is to learn effective policies, typically with access to the global state.
During execution, the goal is to deploy these trained policies in a decentralized, real-world setting without global state information.
The three \ac{MARL} paradigms~\cite{marl-survey} were implemented in the Godot-side to avoid RLlib modifications:

\begin{itemize}
    \item \ac{CTCE} is a single agent solution to the multi-agent problem.
          All observations and actions are concatenated across agents, thus making it a non-scalable solution dependent on a centralized system for execution.
    \item \ac{DTDE}: A fully decentralized approach where each agent trains and executes based on its own local observations.
          This paradigm offers high scalability and robustness~\cite{network_dtde_agents} but often suffers from a lack of coordinated behavior among agents.
    \item \ac{CTDE}: A hybrid paradigm that leverages centralized information during training to learn more coordinated policies, which are executed in a decentralized manner.
    This approach aims to balance the training stability of centralized methods with the scalability of decentralized execution.
\end{itemize}


\subsection{MARL Formulation}
The multi-agent exploration problem is formalized as an \ac{ND-POMDP} to account for partial observability and inter-agent communication.
The framework is defined by the tuple $(\mathcal{N}, \mathbb{S}, \mathbb{A}, T, R, \mathbb{O}, O, \gamma, \mathbb{G})$, where homogeneous agents share identical action and observation spaces ($\mathbb{A}_i = \mathbb{A}_j$, $\mathbb{O}_i = \mathbb{O}_j$ for all $i,j \in \mathcal{N}$).

\textbf{Reward Function}.
The primary objective is to maximize area coverage while avoiding redundant exploration.
The reward function is designed to incentivize continuous discovery:
\begin{equation}
R(s, a, s') = W_{\text{area}} \cdot \Delta Area(s, s'),
\end{equation}

where the coefficient $W_{\text{area}}$ scales the area reward, and
$\Delta Area(s, s')$ represents the new area discovered by the entire team during the state transition, normalized by the theoretical maximum discoverable area $A_{\max}$ per time step $\Delta t$.
This normalization ensures stable learning and prevents gradient explosions.

In detail,
$Area(s)=A_{\text{old}}$ and $Area(s')=A_{\text{new}}$ represent the
total mapped area covered by the team before and after the state
transition, respectively.
The step-wise reward $R_{\text{step}} = R(s,a, s')$ is then calculated based on the normalized newly discovered area, defined as:

\begin{equation}
\Delta Area(s, s') = \frac{Area(s') - Area(s)}{A_{\max}} = \frac{A_{\text{new}} - A_{\text{old}}}{A_{\max}},
\end{equation}

\begin{figure}[htbp]
    \centering
    \includegraphics[width=0.40\textwidth]{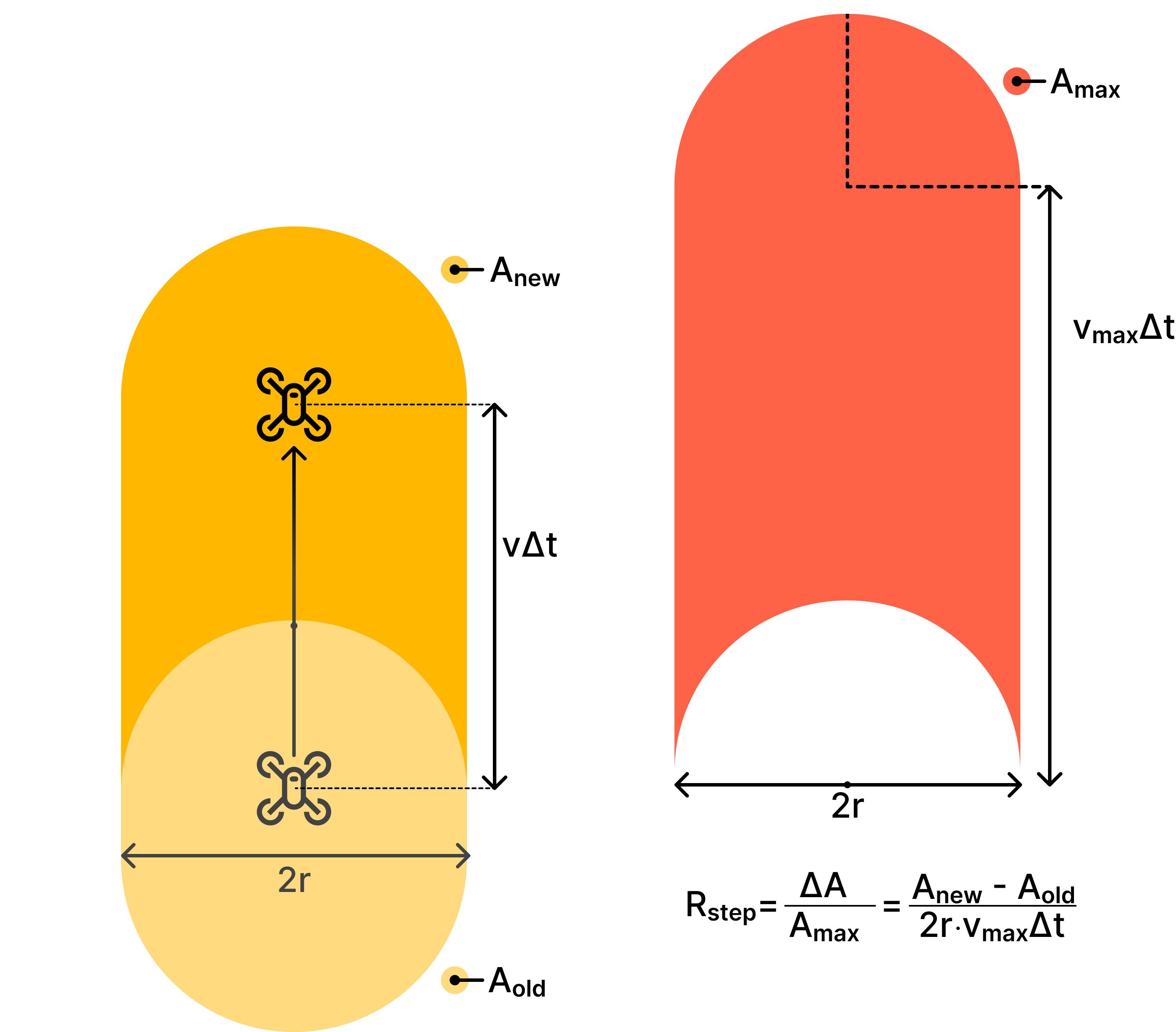}
    \caption{Reward per step.}
    \label{fig:reward_new_area}
\end{figure}

The reward calculation, visualized in~\Cref{fig:reward_new_area}, models sensor coverage as a circular field with radius $r$, equivalent to an ideal \ac{LiDAR} with infinite rays.
The theoretical maximum discoverable area per agent per time step, $A_{\max}$, is calculated based on this coverage model and the maximum
travel distance, given by $A_{\max} = 2r \cdot v_{\max} \Delta t$.
The weighting coefficient $W_{\text{area}}$ enables explicit trade-off calibration against other potential objectives (when used).
For instance, if a collision penalty $R_{\text{collision}} = -1$ is introduced with weight $W_{\text{collision}}$, this quantitative trade-off is defined by the ratio $W_{\text{area}} / W_{\text{collision}}$.
A configuration with $W_{\text{area}} = 1$ and $W_{\text{collision}} = 3$ implies an agent must discover $3$ units of new area to compensate for a single collision.

%
Reward design proved critical in avoiding undesirable emergent behaviors.
Early implementations with collision penalties led to the agents remaining stationary to avoid collisions.
Rewarding only \textit{newly discovered} area, rather than total accumulated area, effectively countered agent passivity and promoted continuous exploration.
Poor exploration strategies are implicitly penalized through opportunity cost (reduced discovered area).

\begin{figure}[htbp]
    \centering
    \includegraphics[width=0.35\textwidth]{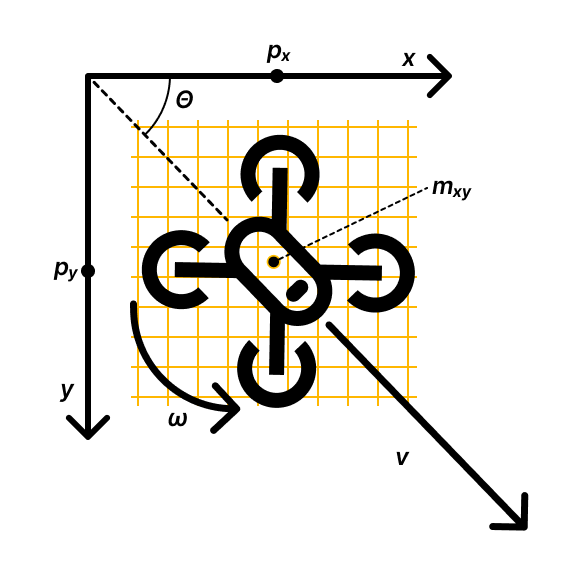}
    \caption{UAV State.}
    \label{fig:state_diagram}
\end{figure}

\textbf{State}.
The system employs homogeneous agents with continuous action spaces to enable realistic \ac{UAV} kinematics.
The state $s \in \mathbb{S}$ comprises the complete 2D map environment, and the individual internal states of the agents characterized by:
\begin{enumerate*}[label=(\roman*)]
    \item position $p =(p_x, p_y)$ using x and y coordinates of the map $m$,
    \item rotation $\theta$,
    \item linear velocity $v =(v_x, v_y)$,
    \item angular velocity $\omega$.
\end{enumerate*}
This is depicted in \Cref{fig:state_diagram}.

\textbf{Actions}.
The action space, defined in \Cref{eq:action_definition}, outputs normalized velocity commands:
\begin{equation}
a_i = (v_x, v_y, \omega) \in [-1, 1]^3
    \label{eq:action_definition}
\end{equation}
These are scaled to realistic \ac{UAV} velocities ($v_{\max} = \SI{400}{px/s}$, $\omega_{\max} = \SI{3}{rad/s}$).
The continuous action space was chosen over discrete alternatives to enable smooth trajectories and avoid exponential dimensionality growth, though it restricts the algorithm choice to policy-gradient methods like \ac{PPO}.

\textbf{Observations}.
Agents perceive their environment through a \ac{LiDAR} sensor.
The position and velocities of the agents are also considered observations.
In a real-world scenario could originate from \ac{GNSS} (outdoor scenarios) and/or \ac{INS}~\cite{drone_gps_ins}.
This work utilizes perfect, noiseless observations from the simulation.
This choice is intentional, as the focus is on \ac{MARL} challenges rather than on perceptual noise robustness--in real world deployments this would be handled by off-the-shelf \ac{SLAM} methods.

Furthermore, the communication protocol can provide agents with access to the \ac{LiDAR} data and Local Maps of other agents.
This information is used to update the agent's own Local Map. 

\textbf{Belief}.
The updated Local Map constitutes the belief of the agent about the environment.
The belief of the agent, implemented as an \ac{OGM}, is updated in one of two ways.
First, both self and shared \ac{LiDAR} scans are incorporated using Bresenham's Line Algorithm to determine occupied and free cells.
Second, shared Local Maps from other agents are fused by summing their log-odds probabilities\footnote{
This is equivalent to a Bayesian update under the assumption of conditional independence. Given cell probabilities $P_1$ and $P_2$ from two maps, the fused probability is $P_c = \frac{P_1 P_2}{P_1 P_2 + (1-P_1)(1-P_2)}$, which corresponds to $L_1 + L_2$ in log-odds space. This assumption is valid as each UAV operates its own sensor.
}.
Because the resulting Local Map is theoretically unbounded, the policy network receives a fixed-size Egocentric Map centered on the agent as its input.

\begin{figure}[htbp]
    \centering
    \includegraphics[width=0.49\textwidth]{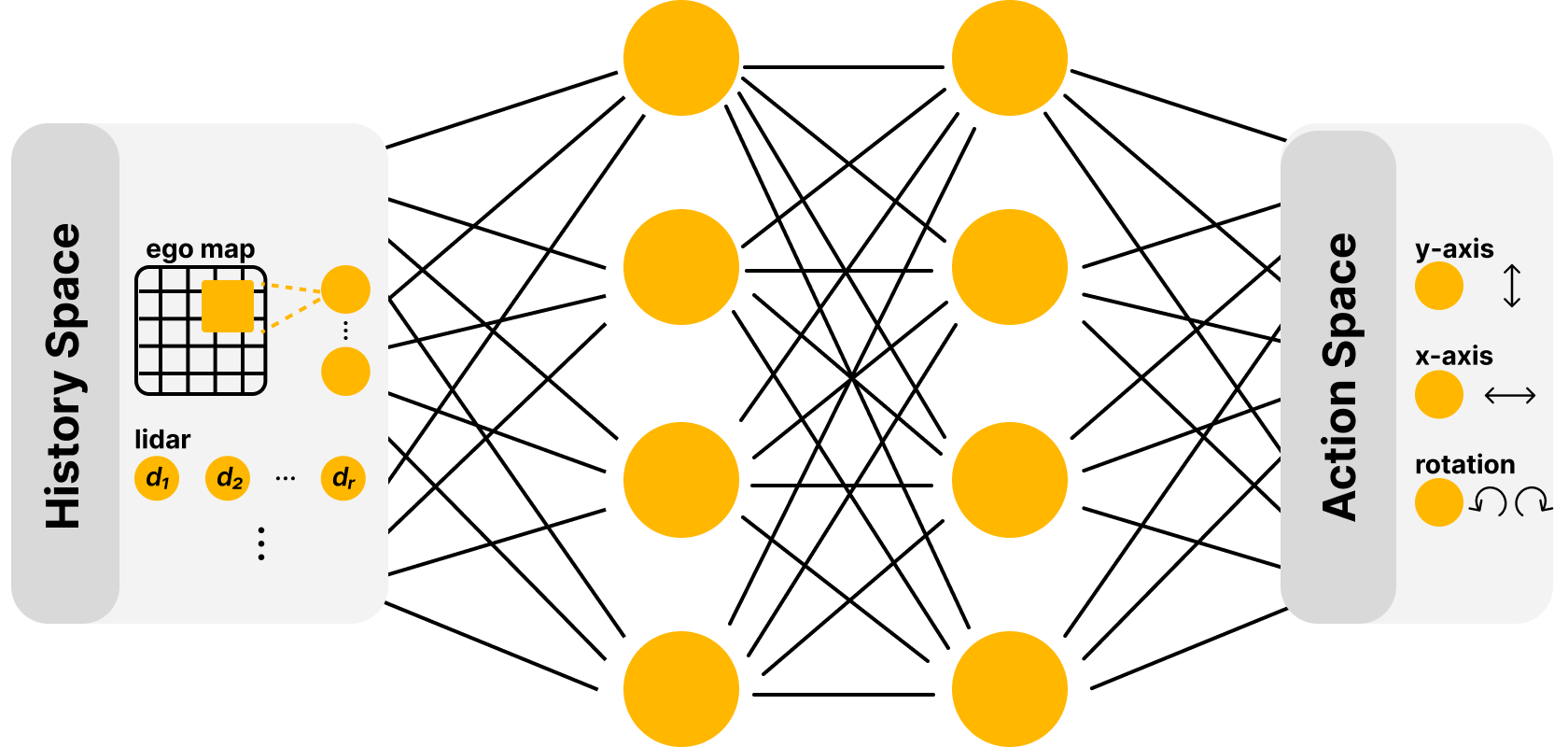}
    \caption{Action Policy Neural Network.}
    \label{fig:agent_action_policy_nn}
\end{figure}

\textbf{Neural Network Architecture}.
Since \ac{PPO} is an Actor-Critic algorithm, each agent employs two neural networks:

\begin{itemize}
\item Action policy ($\pi_a$): Maps the history to continuous actions.
\item Value function ($\pi_v$): Estimates expected cumulative reward for the current state.
\end{itemize}
The Action-policy\footnote{
The value-policy network $\pi_v$ is not depicted, as it shares an identical structure except for the output layer, which consists of a single neuron representing the value estimate of the current state.
} network architecture, illustrated in \Cref{fig:agent_action_policy_nn}, processes the spatially-structured Egocentric Map through a \ac{CNN} while handling the \ac{LiDAR} observation through a Fully-Connected layer.
This hybrid architecture effectively combines spatial reasoning from the map with immediate sensory data from \ac{LiDAR}.
The policy's input and output configurations are modifiable.
For example, the observation space can be extended to include inter-agent distances, and the action space can be reduced to pure 2D translation by excluding rotation.

\begin{figure*}[!hbpt]
\begin{subfigure}[t]{0.139\linewidth}

    \centering
    \includegraphics[width=\linewidth]{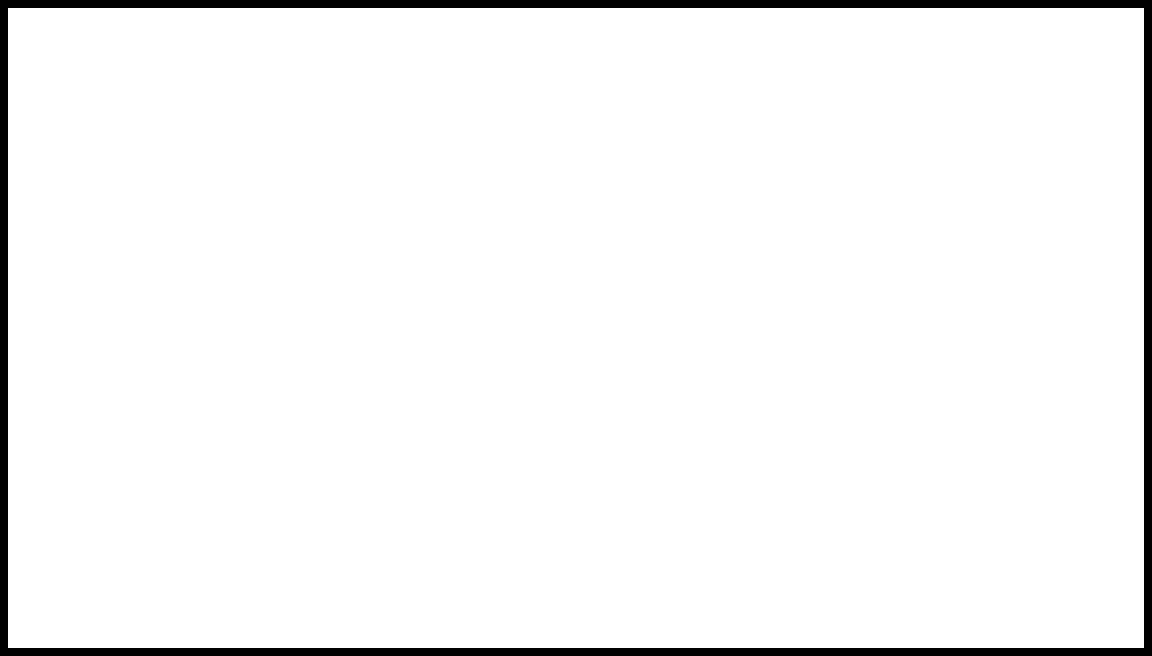}
    \small
    \caption{Level 0}
    \label{fig:level0}

\end{subfigure}\hfill
\begin{subfigure}[t]{0.139\linewidth}

    \centering
    \includegraphics[width=\linewidth]{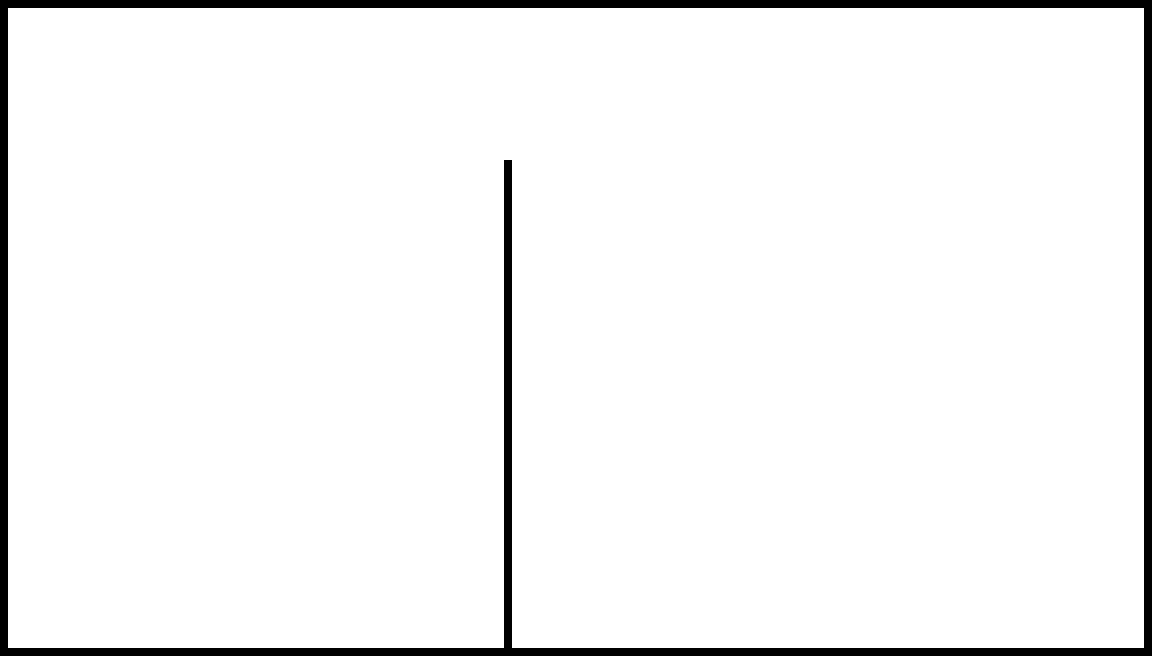}
    \small
    \caption{Level 1}
    \label{fig:level1}

\end{subfigure}\hfill
\begin{subfigure}[t]{0.139\linewidth}

    \centering
    \includegraphics[width=\linewidth]{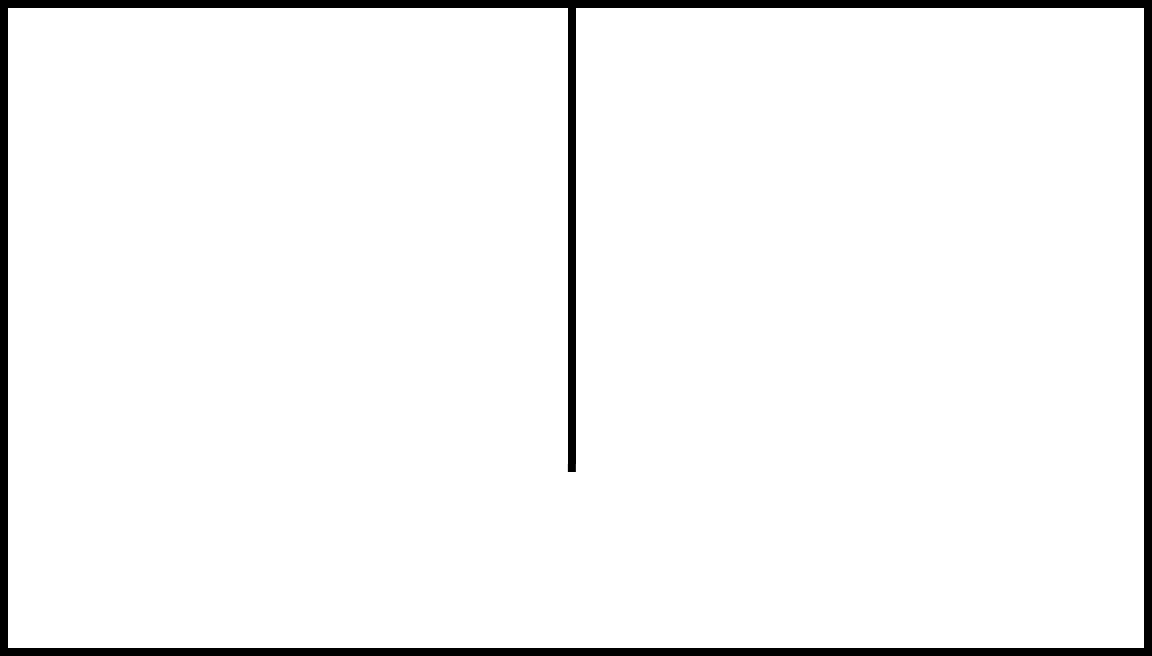}
    \small
    \caption{Level 2}
    \label{fig:level2}

\end{subfigure}\hfill
\begin{subfigure}[t]{0.139\linewidth}

    \centering
    \includegraphics[width=\linewidth]{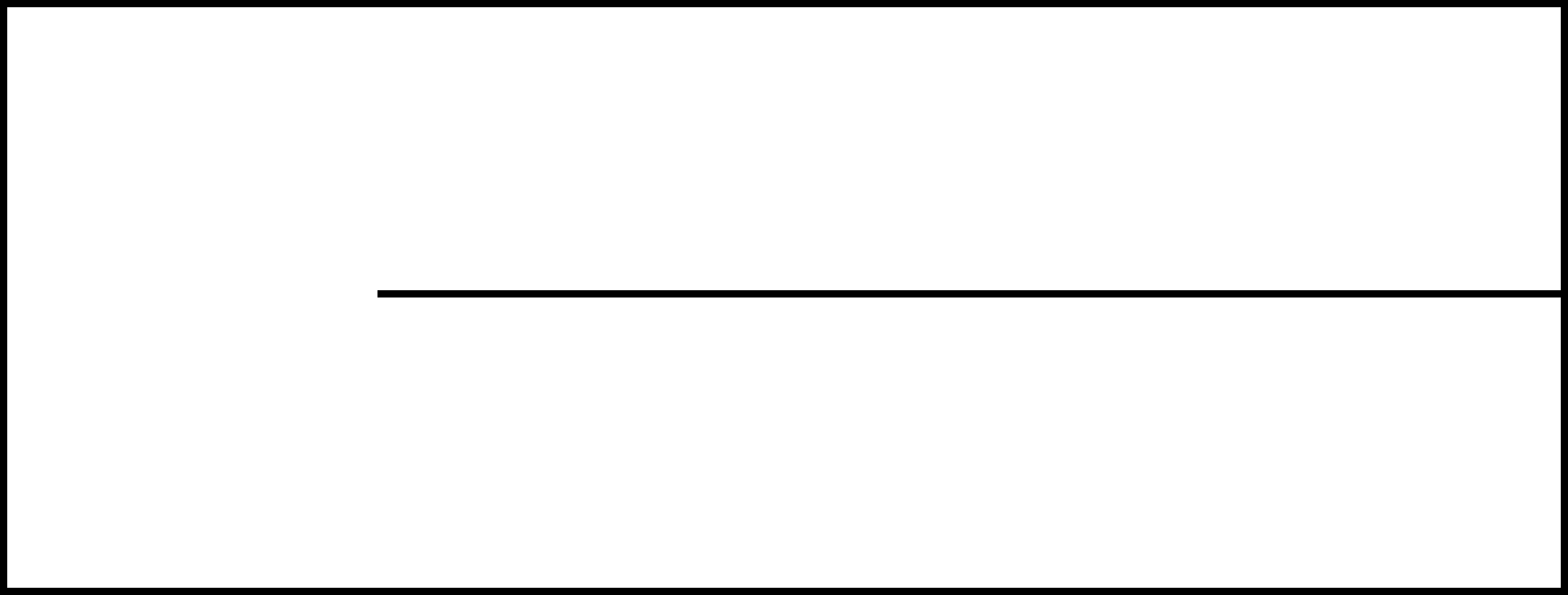}
    \small
    \caption{Level 3}
    \label{fig:level3}

\end{subfigure}\hfill
\begin{subfigure}[t]{0.139\linewidth}

    \centering
    \includegraphics[width=\linewidth]{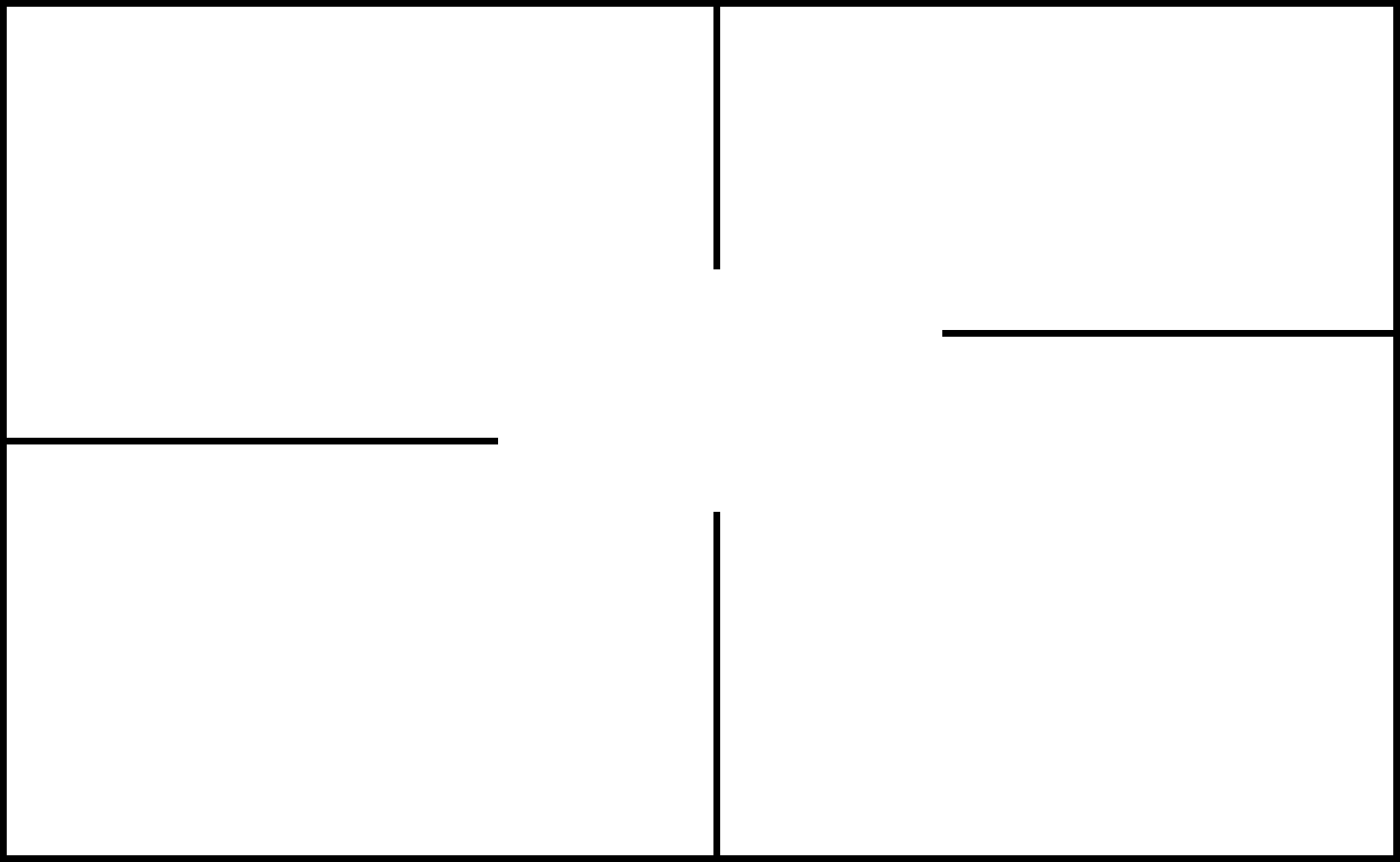}
    \small
    \caption{Level 4}
    \label{fig:level4}

\end{subfigure}\hfill
\begin{subfigure}[t]{0.139\linewidth}

    \centering
    \includegraphics[width=\linewidth]{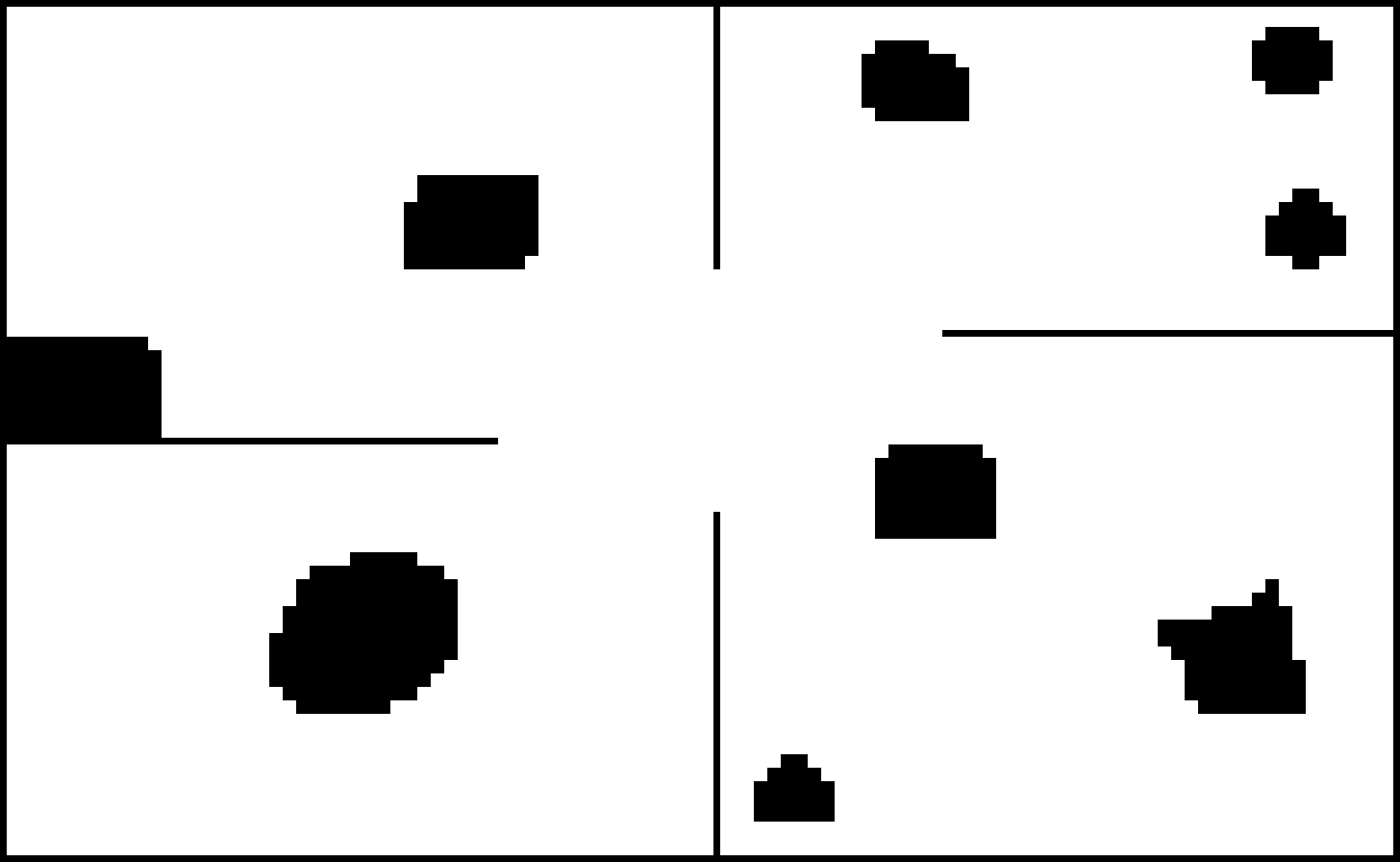}
    \small
    \caption{Level 5}
    \label{fig:level5}

\end{subfigure}\hfill
\begin{subfigure}[t]{0.139\linewidth}

    \centering
    \includegraphics[width=\linewidth]{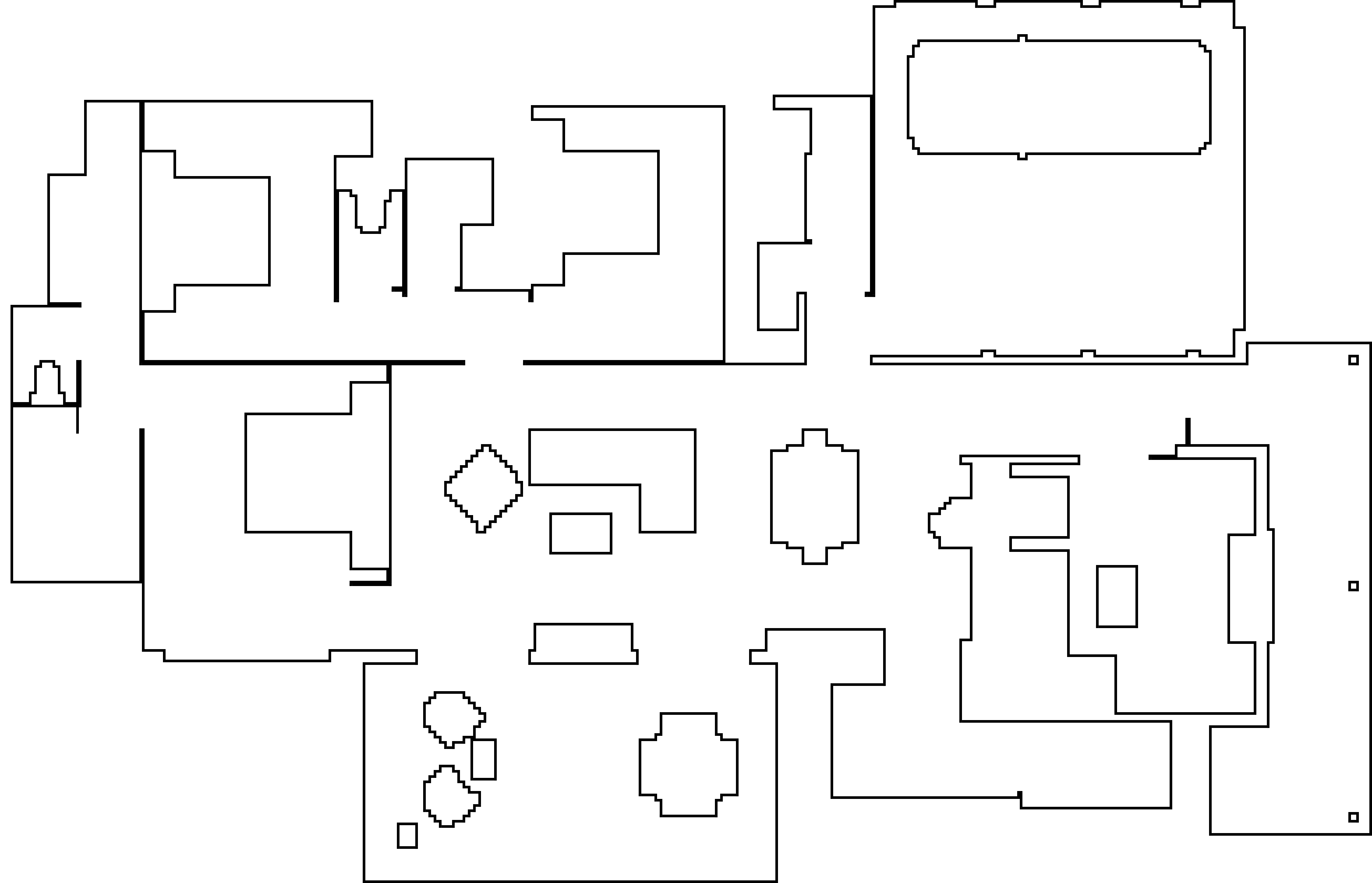}
    \small
    \caption{Level 6}
    \label{fig:level6}

\end{subfigure}
\caption{All Training Levels.}\label{fig:levels_all}
\end{figure*}
\section{Results}
\label{sec:resul}

To account for the inherent stochasticity of \ac{MARL}, this work reports all environment-level metrics as the mean value over multiple episodes.
This practice provides a more stable and robust performance estimate,
with high values indicating that the policy is both high-performing and consistent.

\subsection{Levels}

Agents were tested across seven progressively complex environments
(with increasing area and obstacles/rooms, shown in \Cref{fig:levels_all})
using all combinations of single/multi agent (1, 2, 3 agents), paradigms (\ac{CTCE}, \ac{CTDE}, \ac{DTDE}), and specific levels (0-6).
Each Trial consisting of 1000 Episodes of 1000 Steps.

\begin{figure}[!htbp]
    \centering
\begin{subfigure}[t]{0.7\linewidth}
    \centering
    \includegraphics[width=\textwidth]{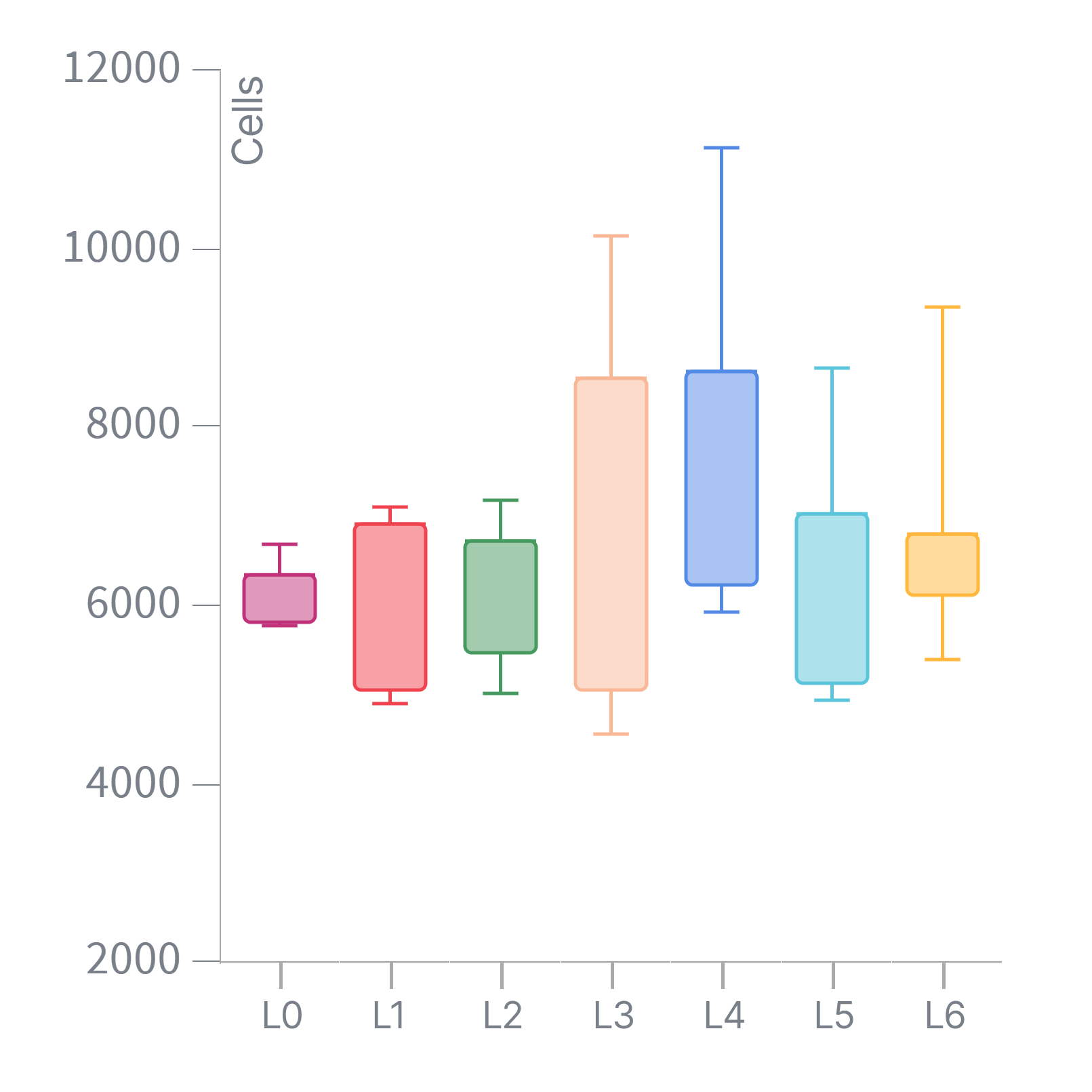}
    \small
    \caption{Progression of level cells explored.}
    \label{fig:box_plot_levels}
\end{subfigure}
\begin{subfigure}[t]{0.7\linewidth}
    \centering
    \includegraphics[width=\linewidth]{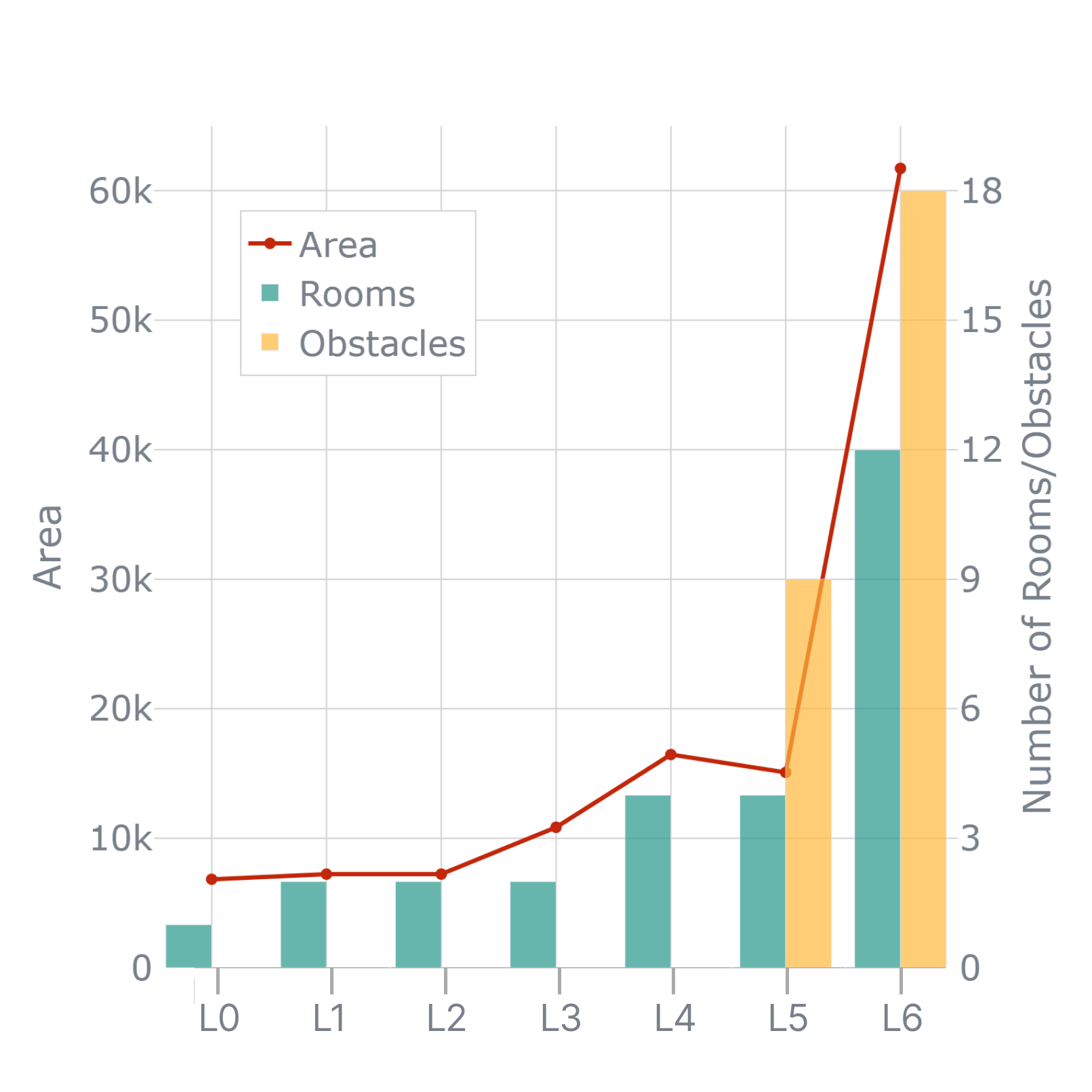}
    \small
    \caption{Progression Difficulty via increasing area and obstacles/rooms.}
    \label{fig:level_difficulty}
\end{subfigure}
    \caption{Comparison of environment exploration and difficulty scaling.}
\end{figure}

All trials showed consistent reward improvement, with final discovered area summarized in \Cref{fig:box_plot_levels}.
Performance increases from L0 to L4 but dips sharply at L5 (8.4\%) due to obstacle-induced navigational challenges, before rising again at L6.
This progression in difficulty is further illustrated in \Cref{fig:level_difficulty}.

\begin{figure}[!]
    \centering
\begin{subfigure}[t]{1.0\linewidth}
    \centering
    \includegraphics[width=\linewidth]{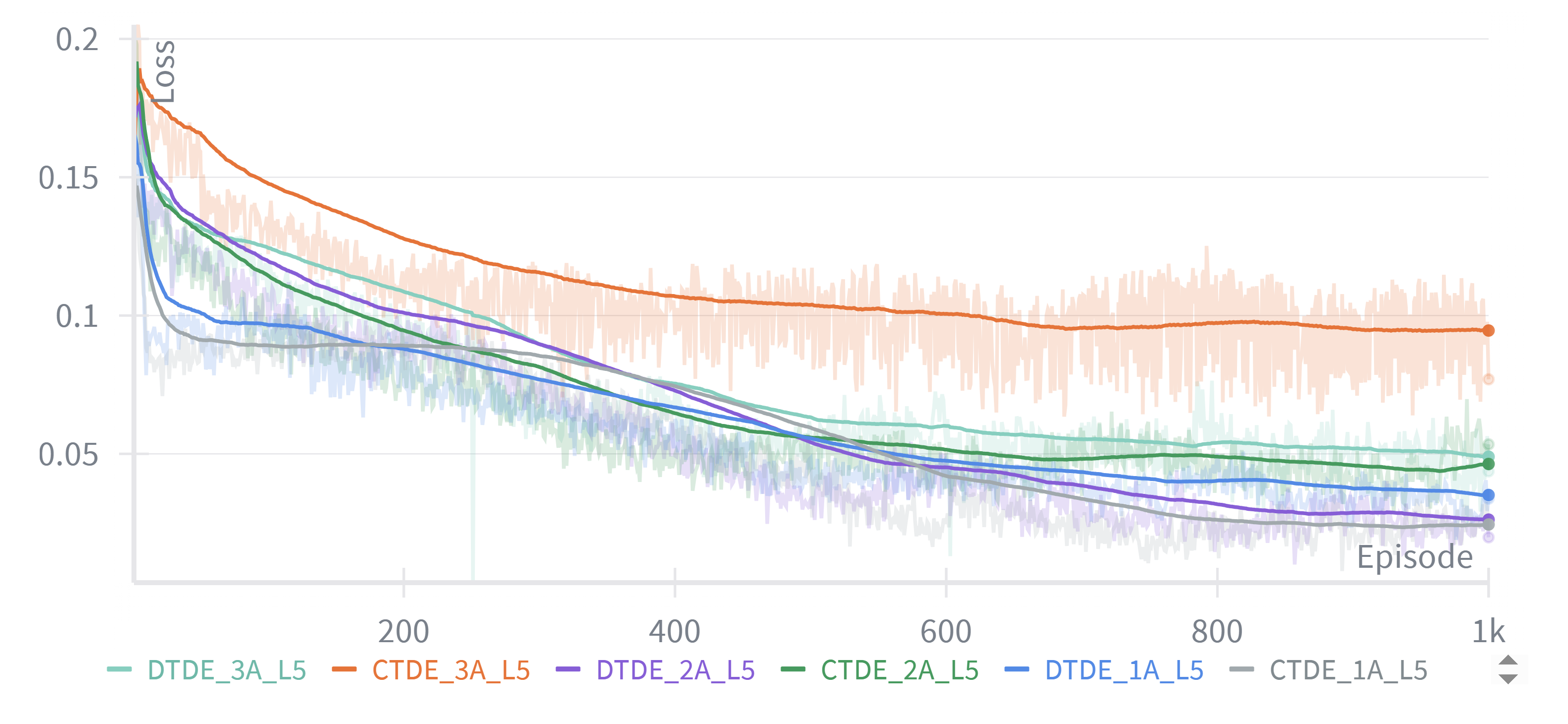}
    \small
    \caption{Policy Loss}
    \label{fig:policy_loss}
\end{subfigure}
\begin{subfigure}[t]{1.0\linewidth}
    \centering
    \includegraphics[width=\linewidth]{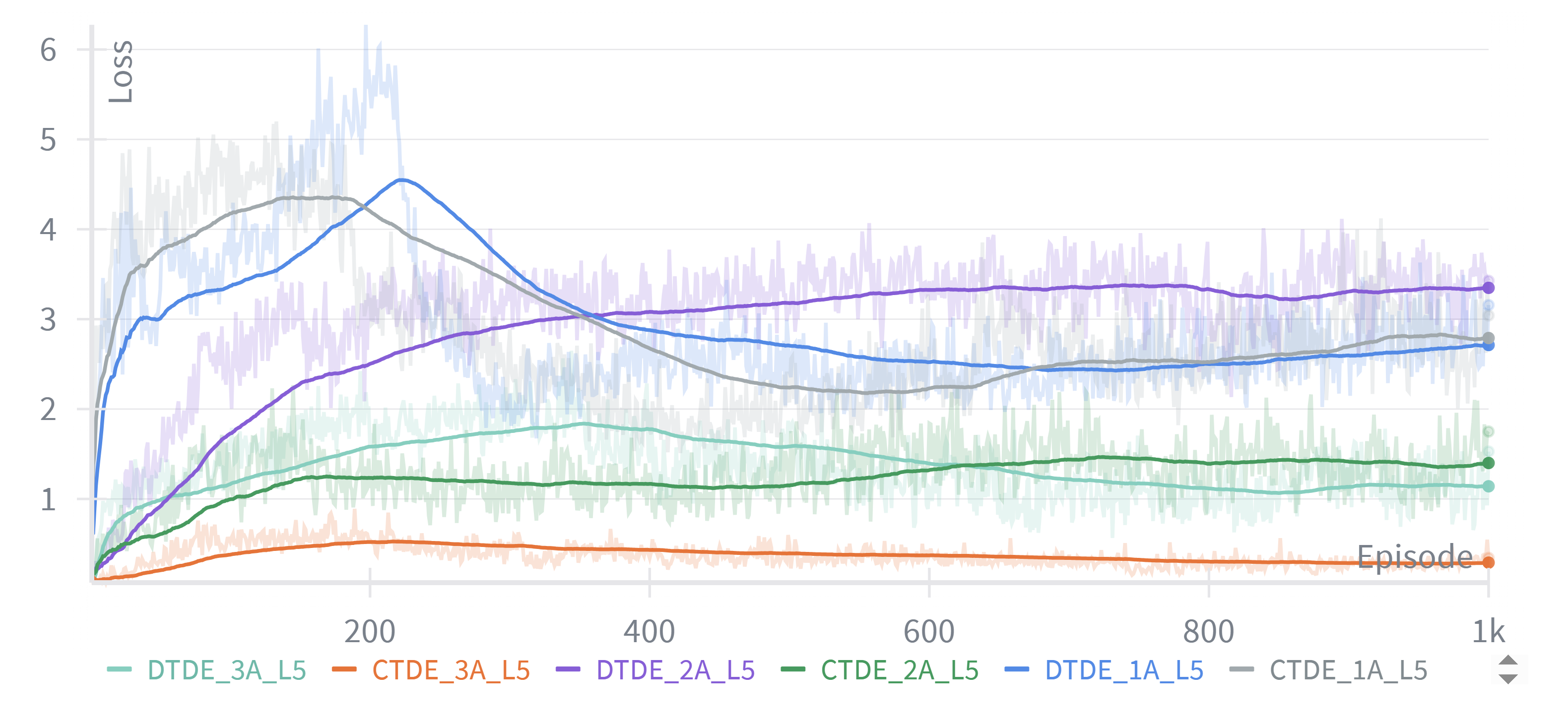}
    \small
    \caption{Value Function Loss}
    \label{fig:vf_loss}
\end{subfigure}
    \caption{Policy and Value Function Loss.}
\end{figure}

The policy loss demonstrates a consistent decrease, indicating continuous improvement throughout training, in~\Cref{fig:policy_loss}.
Additionally, the value function loss exhibits the expected pattern\footnote{
An initial rise followed by a decline.
This occurs because as agents begin to learn and receive higher rewards, the value function must adapt
to these new, higher value estimates, initially increasing the loss. Once learning stabilizes and rewards
become more predictable, the value function converges to accurate state-value estimates, resulting in a
decreasing loss.} shown for Level 5 in~\Cref{fig:vf_loss}, and is representative of all Levels.
For improved clarity, \Cref{fig:policy_loss,fig:vf_loss} utilize \ac{EWMA}~\cite{ewma}.


\pagebreak
Key findings include the following:
\begin{itemize}
\item Scalability: Multi-agent configurations consistently outperformed single-agent setups, achieving up to 98\% coverage in simpler levels (\Cref{fig:level0,fig:level1,fig:level2}).
\item Multi-Agent Complexity: 2-agent configurations occasionally outperformed 3-agent systems, demonstrating the that the coordination complexity increases non-linearly with team size.
\item Paradigms: \ac{CTCE}, as a single-agent paradigm, greatly simplifies the problem, which accounts for its strong performance.
However, \ac{CTDE} emerged as the most effective training paradigm, particularly in complex environments where centralized value estimation provided advantages over the fully decentralized approach \ac{DTDE}.
\item Environmental Complexity: Performance naturally decreased with increasing obstacle density and exploration area, though absolute exploration capability improved as agents discovered more cells in larger environments.
\end{itemize}

\begin{figure*}[!t]
    \centering
\begin{subfigure}[t]{0.33\linewidth}

    \centering
    \includegraphics[width=\linewidth]{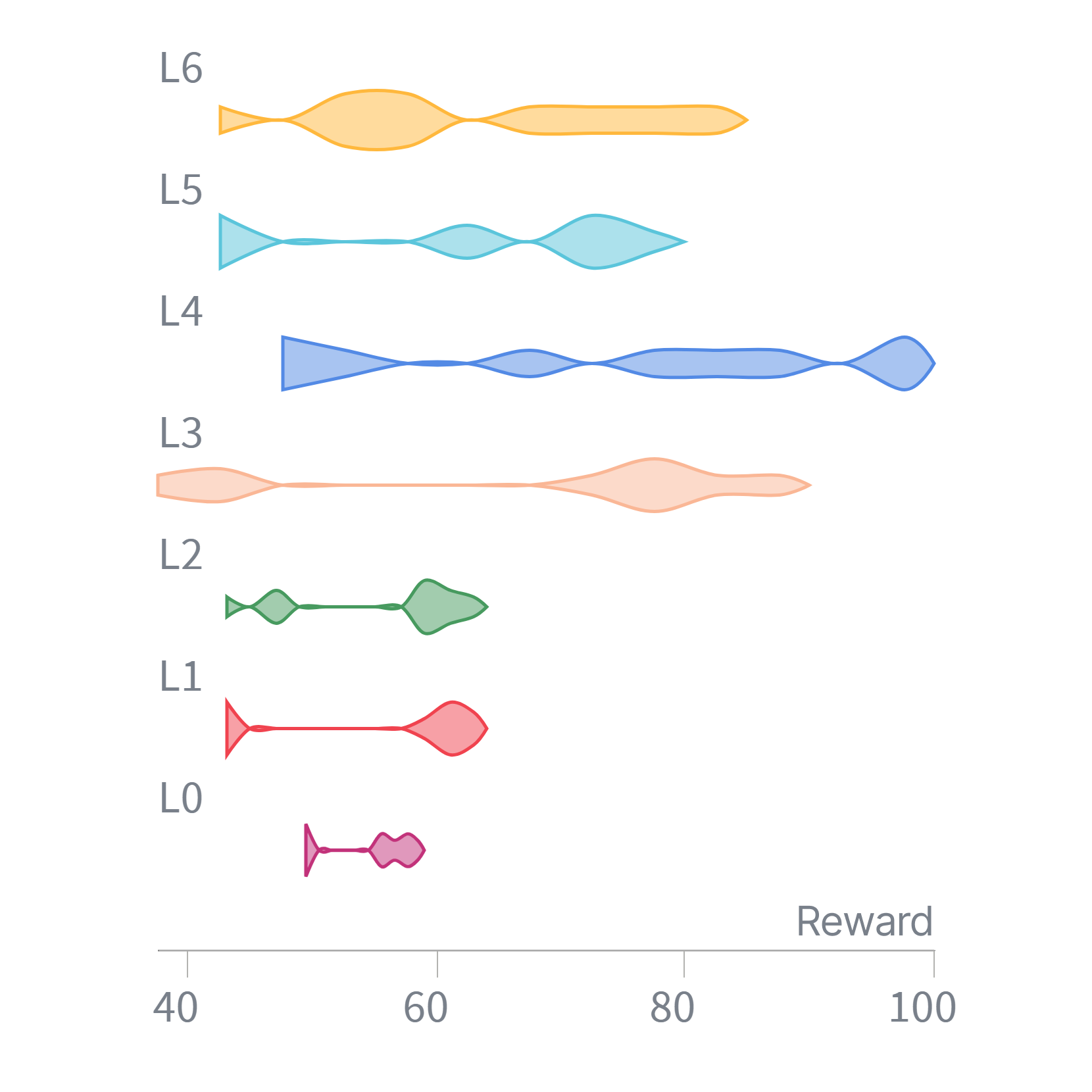}
    \small
    \caption{Level Reward.}
    \label{fig:batch_reward_per_level}

\end{subfigure}
\begin{subfigure}[t]{0.33\linewidth}

    \centering
    \includegraphics[width=\linewidth]{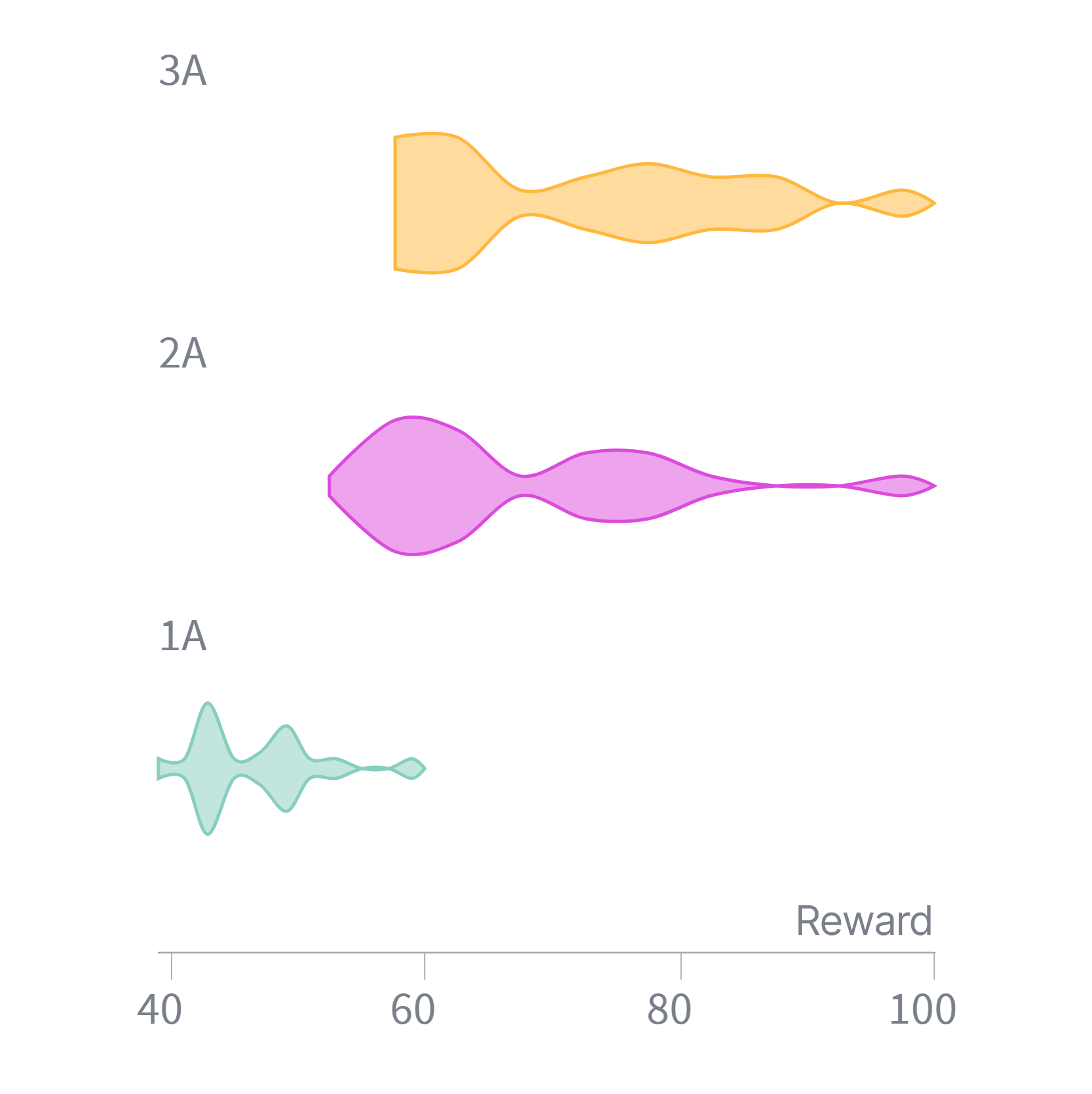}
    \small
    \caption{Paradigm Reward.}
    \label{fig:batch_reward_per_paradigm}

\end{subfigure}
\begin{subfigure}[t]{0.33\linewidth}

    \centering
    \includegraphics[width=\linewidth]{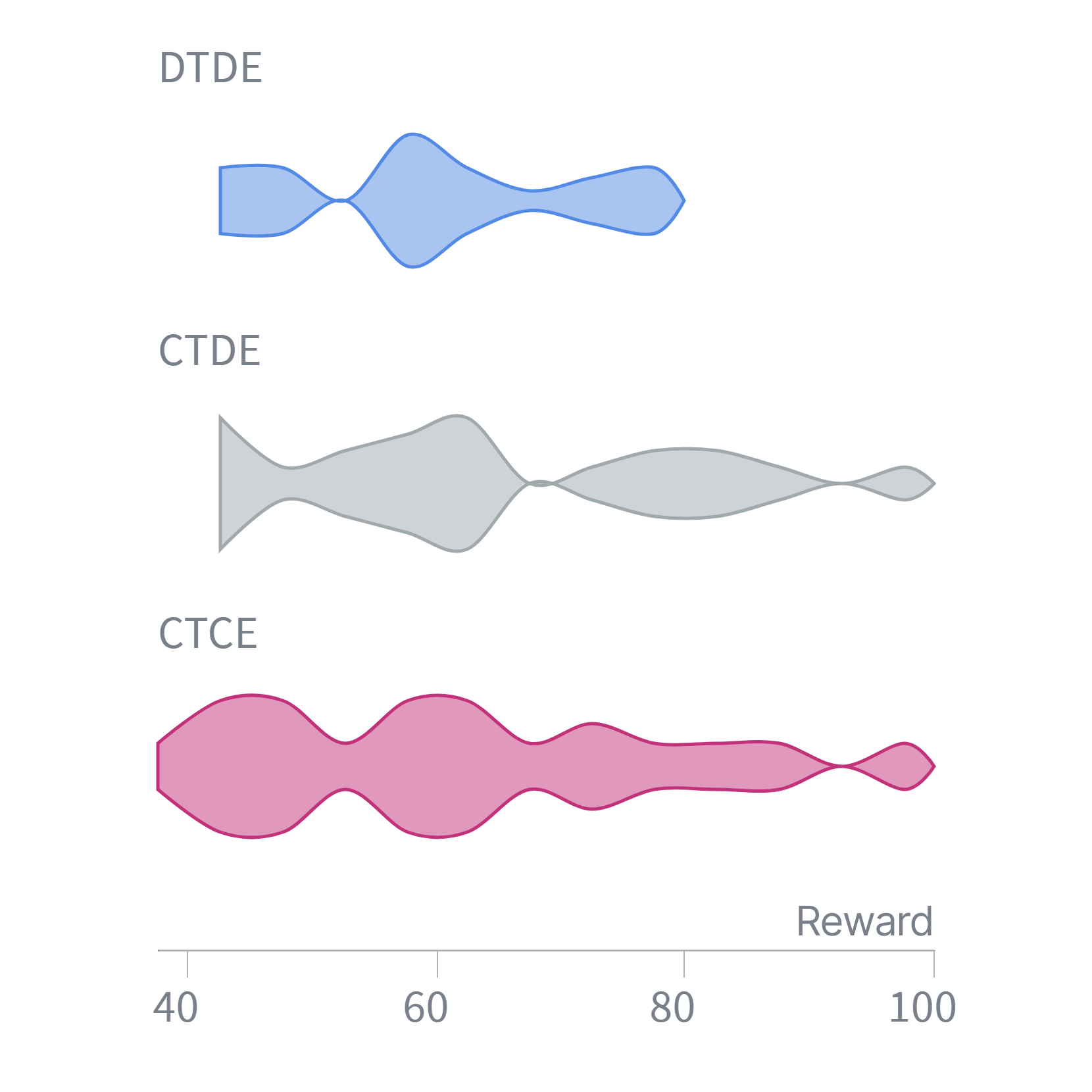}
    \small
    \caption{Number of Agents Reward.}
    \label{fig:batch_reward_per_agent}

\end{subfigure}
\caption{Reward per Level, Paradigm and Number of Agents.}
\end{figure*}
Cross-level analysis (\Cref{fig:batch_reward_per_level,fig:batch_reward_per_paradigm,fig:batch_reward_per_agent}) confirm that higher agent counts yield superior results,
and that the higher levels are more challenging, because of the gradual increase in reward range.
Note that the higher levels are also larger, meaning a good
navigating agent would gain a larger reward in the harder levels.
This is evident for Level 6, the largest
and most demanding, which has a large reward range with an even distribution.

\subsection{Communication and Coordination}

\begin{figure}[!htbp]
\begin{subfigure}[t]{1.0\linewidth}
    \centering
    \includegraphics[width=\linewidth]{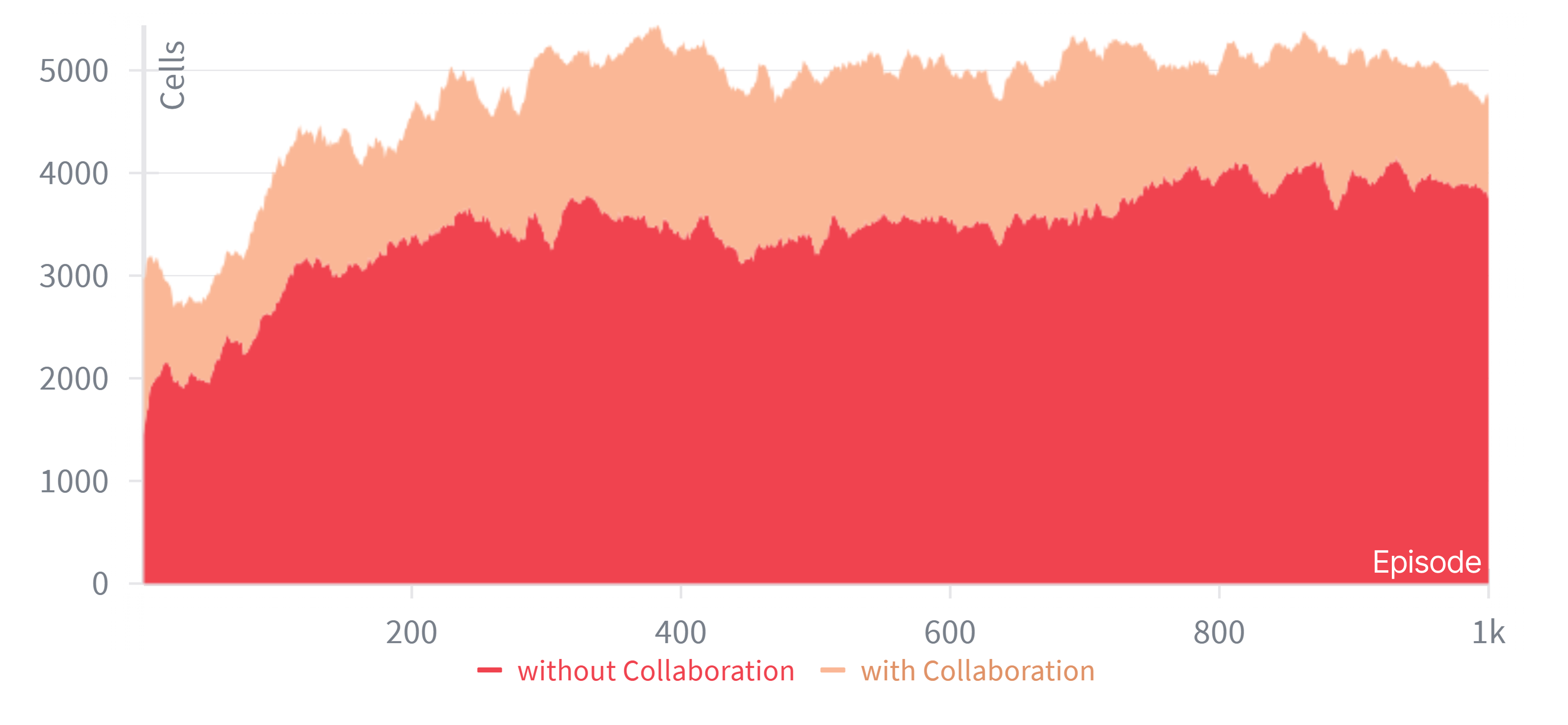}
    \small
    \caption{Cells acquired through Discovery and Collaboration.}
    \label{fig:discovered_cells}
\end{subfigure}
\begin{subfigure}[t]{1.0\linewidth}
    \centering
    \includegraphics[width=\linewidth]{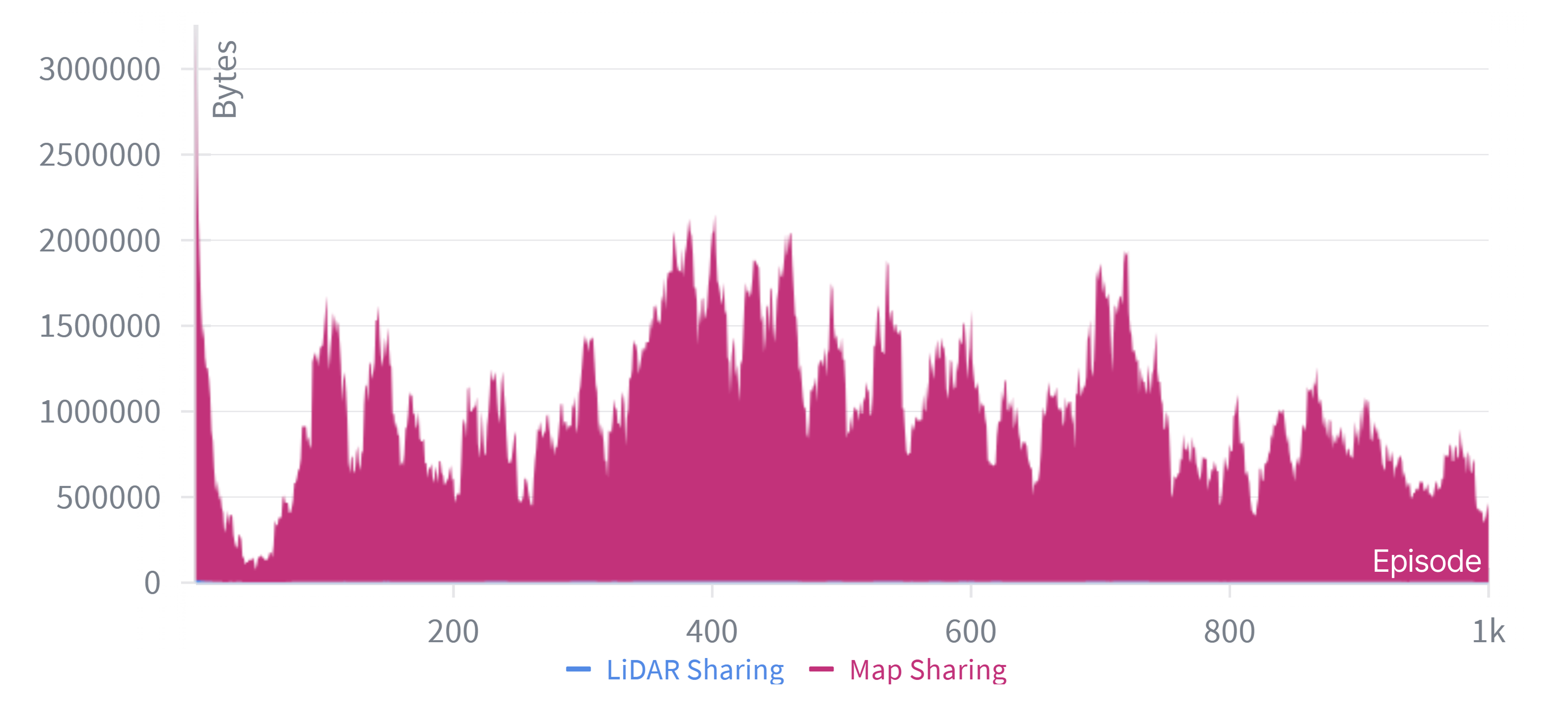}
    \small
    \caption{Bytes Transmitted in Collaboration.}
    \label{fig:bytes_sent}
\end{subfigure}
\caption{Cells from Discovery and Collaboration, and the Bytes Transmitted.}
\end{figure}

Analysis of inter-agent communication revealed significant collaborative benefits:

\begin{itemize}
\item Agents effectively shared environmental information, with communication accounting for up to 40\% of discovered cells (as shown in \Cref{fig:discovered_cells}).
\item \ac{LiDAR} Sharing proved more effective (responsible for 85\% of shared cells declining to 60\% over time) and more efficient than full Map Sharing\footnote{
This inefficiency stems from transmitting the entire map rather than only discovered portions,
and the absence of compression techniques such as Run-Length Encoding or LZMA2 (e.g., as used in 7zip).
} (1\% vs 99\% of the bandwidth, as shown in \Cref{fig:bytes_sent}),
though both strategies contributed to exploration.
\item Hypothetical emergent behaviors include both continuous motion coordination (maintaining proximity for \ac{LiDAR} Sharing) or periodic regrouping for map synchronization.
\item Cross-level analysis reveals bounded exploration and sharing dynamics: the ratios between agents remain within $\interval{0.5}{2}$.
This indicates that at any given time, one agent discovers at most twice as many cells as another, and similarly shares at most twice as many new cells.
\end{itemize}

To further quantify the effects of cooperation,
team sizes of 2–4 agents were tested under the CTDE paradigm across three communication protocols:
no communication (\textit{off}), a constrained one-hop protocol (\textit{one-hop}),
and a multi-hop network (\textit{multi-hop}) where messages are rebroadcast by all agents.

\begin{figure}[!t]
    \centering
    \includegraphics[width=0.49\textwidth]{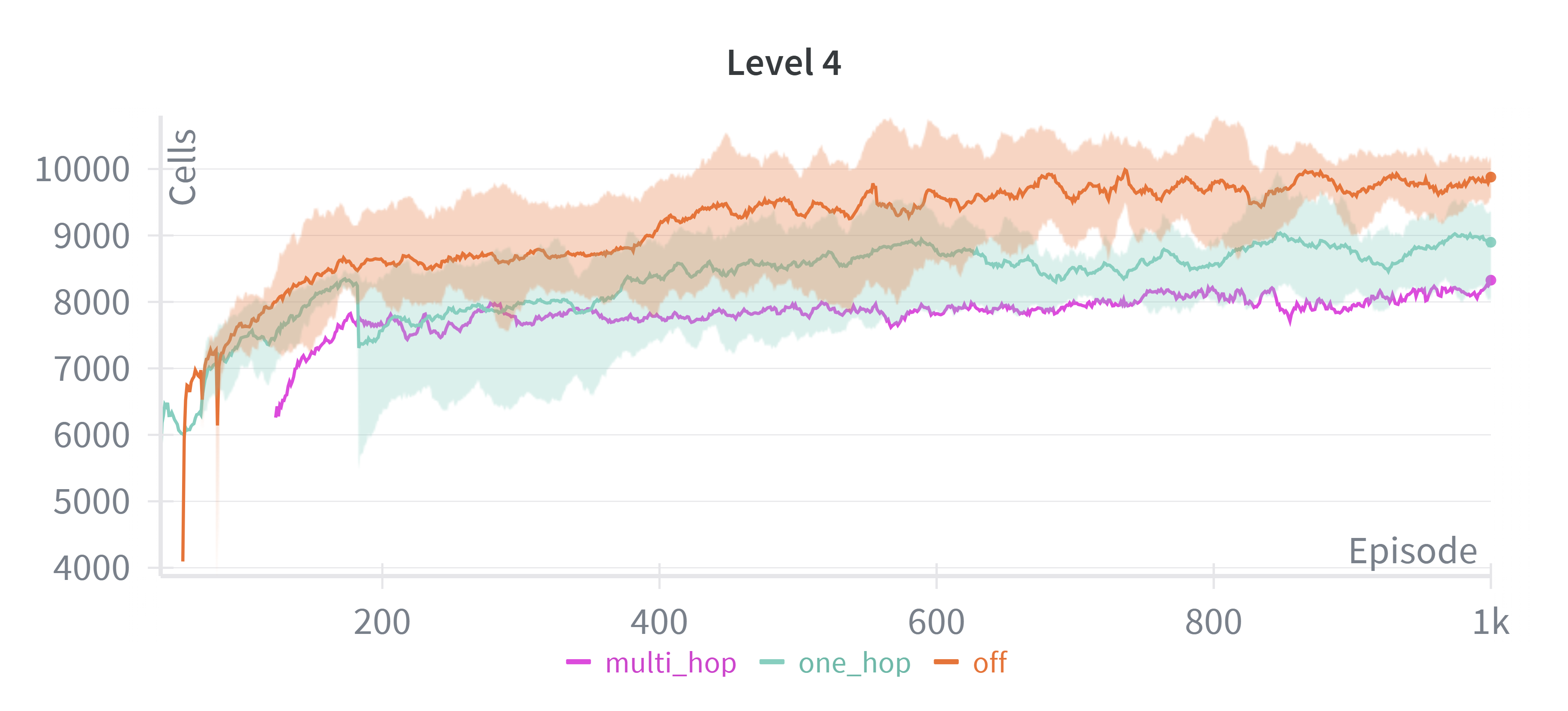}
    \caption{Performance across different communication protocols.}
    \label{fig:comm_paradox}
\end{figure}

The results, presented in \Cref{fig:comm_paradox}, revealed a counterintuitive pattern: increased communication leads to worse overall performance.


It is hypothesized that this paradox arises because agents, through shared map information, learn to avoid areas marked as explored.
In Level 4--which features a central chamber that must be traversed to access other rooms--this behavior becomes detrimental.
If multiple agents learn to avoid this central area after initial exploration, they become trapped and unable to reach larger unexplored regions beyond it.

Thus, reducing communication limits shared map data, inadvertently helping agents identify unexplored areas.
This further indicates that the current policy architecture lacks complexity required for navigating through a known area to reach unexplored regions.

\subsection{Ablation Studies}

\textbf{Kill on Collision}.
This paper investigated variable-length episodes with early termination upon agent collisions, termed \textit{Kill on Collision},
with the hypothesis that this would incentivize collision avoidance and improve navigation performance.
Contrary to expectations, fixed-length episodes substantially outperformed this approach by a factor of 5.

Building on insights from single-agent formulations~\cite{learning_cp_unkown_env_drl}, a collision penalty $R_{\mathrm{collision}} = -3$ was introduced.
However, this resulted in the agents learning to become immobilized to avoid potential collisions. 

These findings highlight that \ac{MARL} formulations present substantially greater complexity than single-agent scenarios, requiring careful consideration of reward structures and termination conditions to avoid unintended behaviors.

\textbf{Parallel vs Sequential}.
Given the hierarchical structure of the environment levels\footnote{
The levels monotonically increase in explorable area and obstacle density.
Level 1 introduces basic walls, Level 3 adds
obstacles, Level 4 expands the scale with more rooms, Level 5 increases obstacle density, and Level 6
serves as the most realistic and complex scenario, created using a house blueprint.
}, two Curriculum-Learning strategies were evaluated:
\begin{enumerate*}[label=(\roman*)]
    \item Parallel training on all levels simultaneously, and
    \item Sequential training, where agents progress through levels one-by-one (upon reaching 80\% area coverage)
\end{enumerate*}.

Sequential progression proved more effective, achieving a 250\% higher performance than the Parallel strategy.
this result is attributed to \textit{catastrophic interference} in the Parallel setting,
where the diverse and conflicting dynamics from all levels simultaneously overwhelm the policies of the agents, preventing stable learning.
In contrast, the Sequential curriculum allows the agent to construct knowledge incrementally,
mastering fundamental skills in earlier levels that robustly transfer to more complex ones.

\textbf{Curriculum-Learning}.
A key limitation of the previous Sequential training was the overly simplistic passing condition.
Requiring 80\% area coverage to be achieved only once could result from policy exploration rather than demonstrating a consistent ability.

Parametrizing via \texttt{pass\_area} (coverage threshold) and \texttt{pass\_x\_times} (consistency requirement), for example, reaching 80\% area 20 times, allowed this
Curriculum-Learning formulation to enhance performance, and speed up training.

\begin{figure}[htbp]
    \centering
    \includegraphics[width=0.48\textwidth]{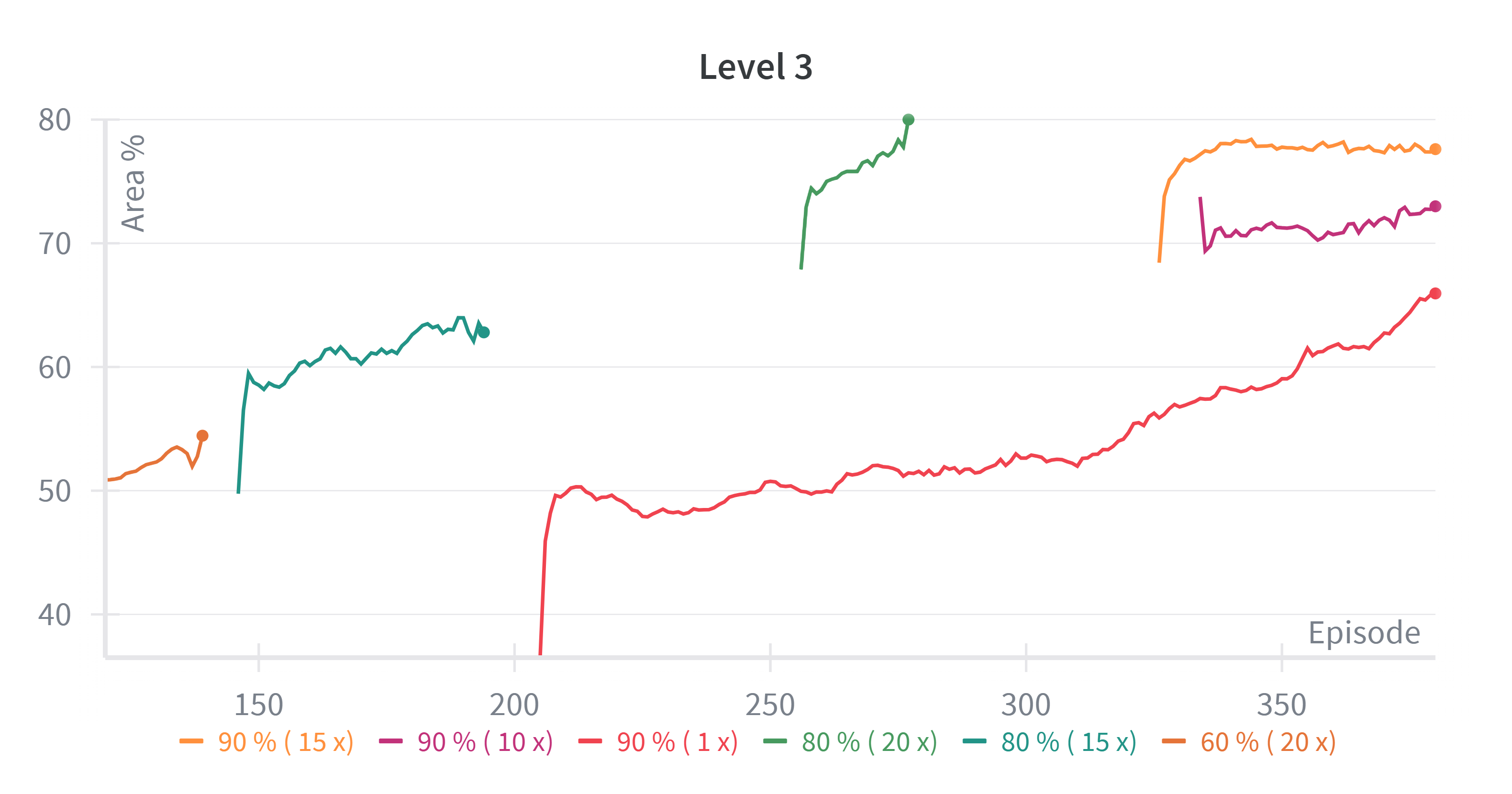}
    \caption{Level 3.}
    \label{fig:curriculum_level_3}
\end{figure}

\Cref{fig:curriculum_level_3} illustrates \texttt{pass\_area} varying
between 60, 70, 80 and 90\% and \texttt{pass\_x\_times} varying between 1, 10, 15 and 20.
Level 3 revealed clear interactions between threshold difficulty and required consistency.

\Cref{fig:curriculum_level_3} shows varying intervals resulting from changing passing conditions.
Several key behaviours were identified:
\begin{enumerate}[label=(\arabic*)]
    \item Harder progression criteria enhance learning,
    the \textcolor[HTML]{479a5f}{80\% (20x)} demonstrates immediate performance advantages over \textcolor[HTML]{229487}{80\% (15x)}, which in turn outperforms \textcolor[HTML]{e57439}{60\% (20x)}.
    likewise the 90\% \textcolor[HTML]{ff9c3d}{(15x)}$>$\textcolor[HTML]{c2327a}{(10x)}$>$\textcolor[HTML]{f0434f}{(1x)}, confirming a positive correlation between progression difficulty and learning effectiveness.

    \item Reliability Over Early Success:
    The \textcolor[HTML]{f0434f}{90\% (1x)} condition advances to Level 3 approximately 50 episodes earlier than \textcolor[HTML]{479a5f}{80\% (20x)}, yet the latter achieves immediately superior performance.
    The single-success condition shows only gradual improvement and remains consistently dominated by multi-success variants (\textcolor[HTML]{ff9c3d}{15x}, \textcolor[HTML]{c2327a}{10x}), validating the emphasis on \textit{reliable} performance over episodic luck.

    \item Perfectionism:
    Excessively high coverage thresholds (90\%) with variants \textcolor[HTML]{ff9c3d}{(15x)}, \textcolor[HTML]{c2327a}{(10x)}, and \textcolor[HTML]{f0434f}{(1x)} achieve lower overall performance than \textcolor[HTML]{479a5f}{80\% (20x)}.
    Reaching the higher 90-100\% coverage rates proves very challenging, because of the mapping gaps during movement
    and the possible need for backtracking through discovered area to locate small remaining gaps.

    \item Balance:
    The \textcolor[HTML]{479a5f}{80\% (20x)} condition achieves an area mean of 80\%, suggesting that these hyperparameters may represent an optimal balance, enforcing consistent performance without excessive perfectionism.
\end{enumerate}


\subsection{Optimizing Actions and Observations}
The policies of the agents exhibited significant sensitivity to the formulation of both actions and observations.
A grid search over these design possibilities revealed clear performance hierarchies.

When grouping by action space, a consistent ranking emerged: movement in a 2D plane without rotation outperformed 1D movement with independent rotation, both of which substantially surpassed a waypoint-based approach where the agent specifies a target direction and the drone automatically rotates and moves\footnote{Manual control in this mode proved challenging and unintuitive, warranting further refinement. Such control mechanisms resemble two-stage RL formulations, introducing additional learning complexity that conflicts with the end-to-end approach utilized by this work.}.

Analysis by observation space showed less pronounced differentiation, though trendlines consistently favored the combination of the Egocentric Map with \ac{LiDAR} data.
The explicit inclusion of inter-agent distances proved detrimental to learning efficiency.
It is hypothesized that \ac{LiDAR} measurements already implicitly encode proximity information, albeit without distinguishing between agent and obstacle contacts.
Meanwhile, \ac{LiDAR}-only observation proved insufficient, aligning with expectations since only the Egocentric Map incorporates historical observations, providing the essential temporal context for effective navigation.


\subsection{Architectural Optimizations}
\begin{figure}[!b]
\centering
\begin{subfigure}[t]{0.48\textwidth}
    \centering
    \includegraphics[width=\textwidth]{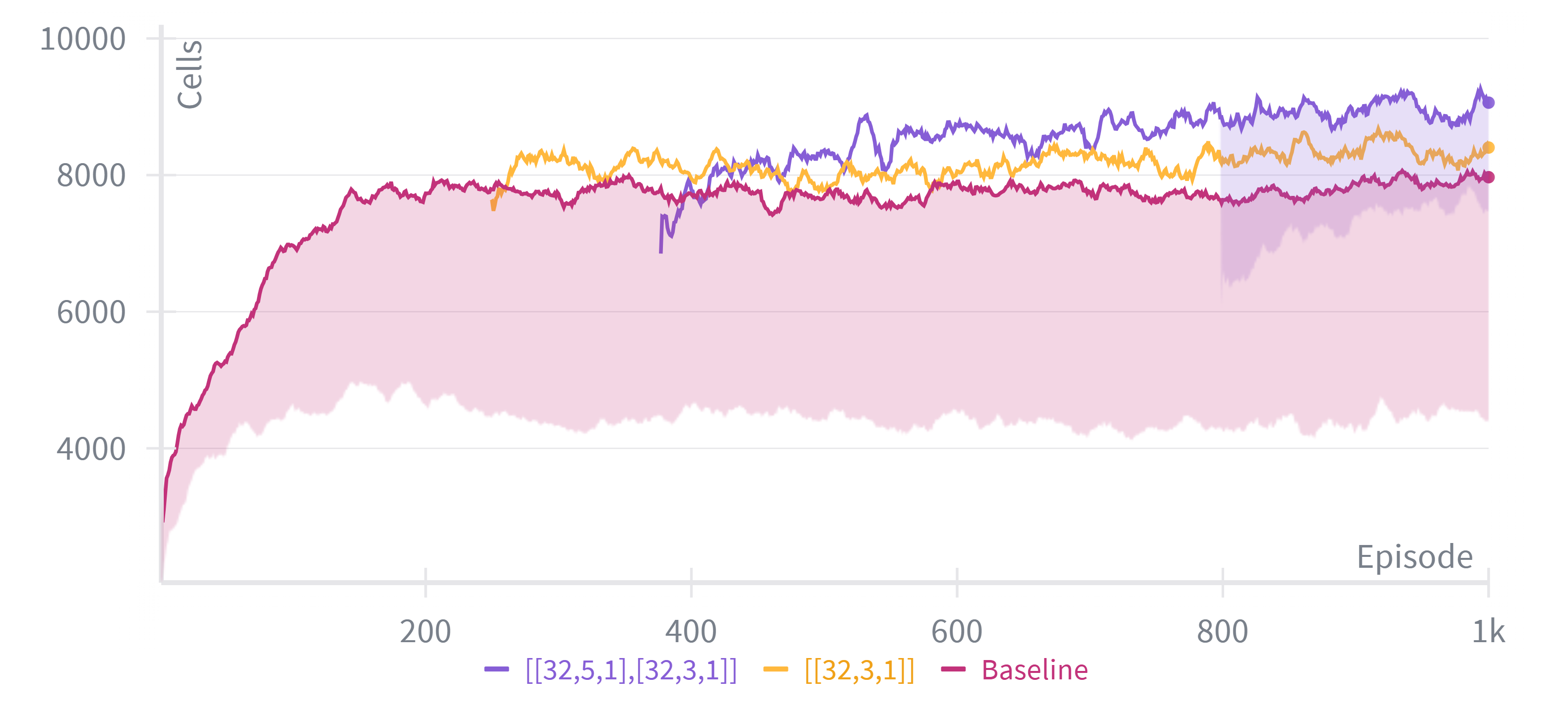}
    \caption{LiDAR Convolutions.}
    \label{fig:level5_lidar_conv}
\end{subfigure}
\vspace{0.5em}
\begin{subfigure}[t]{0.48\textwidth}
    \centering
    \includegraphics[width=\textwidth]{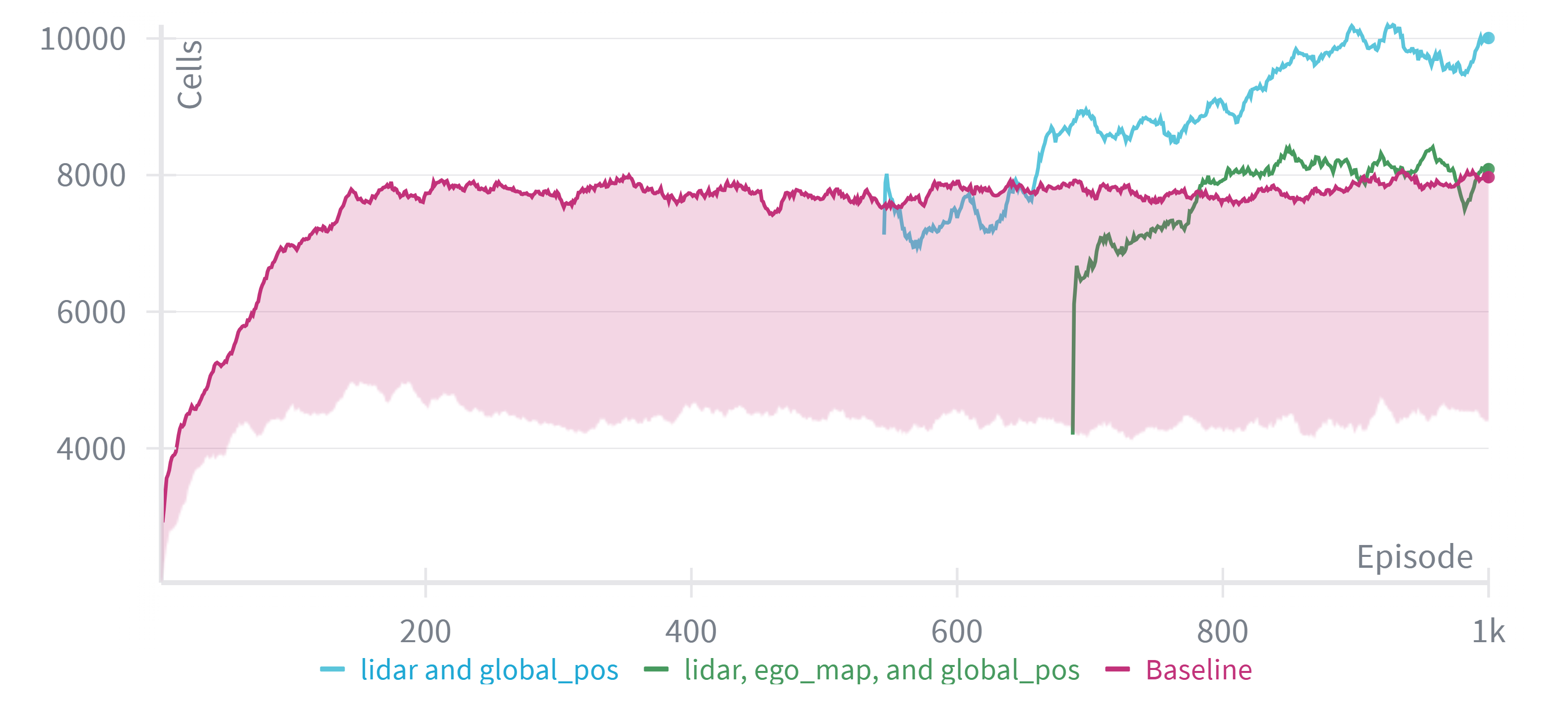}
    \caption{RNNs.}
    \label{fig:level5_rnn}
\end{subfigure}
    \caption{Architectural optimizations in Level 5: (a) convolutional processing of LiDAR data or (b) the integration of RNNs for temporal reasoning.}
\end{figure}
Several neural network enhancements significantly improved performance.

\noindent
\textbf{Centralized Critic}.
Parameter sharing across value functions provided a small improvement ($1$\% area coverage).

\noindent
\textbf{LiDAR Convolutions}.
1D \ac{CNN} processing of \ac{LiDAR} data increased performance by 20\%, allowing agents to overcome previous performance barriers and advance to the more complex Level 5 (\Cref{fig:level5_lidar_conv}).
Two different convolution filters were tested: a single and a double layer one, with the deeper layered one showing better performance.

\noindent
\textbf{RNN Integration}.
Temporal processing improved performance (13\%), with optimal configurations combining \ac{LiDAR} and global positioning data (\Cref{fig:level5_rnn}), though at the cost of slower training convergence.
The baseline achieved comparable cell discovery within 70 episodes, while the \ac{RNN} approach required 220 episodes to surpass this threshold.

The best performing configurations used global positioning data.
This enables agents to learn panning motions for systematic area coverage by leveraging temporal sequences of global positions.
Furthermore, by analyzing position history, agents can reason about unexplored regions, raising questions about global versus local planning strategies.
The current formulation optimizes for local exploration by incentivizing immediate area discovery, whereas a global approach might initially target less rewarding areas to minimize future backtracking.
Global planning approaches are more aligned with the objective of achieving complete area coverage (90-100\%),
which represents a related problem, but beyond the scope of this work.


The inferior performance of \acp{RNN} with only \ac{LiDAR} observations,
compared to the belief-based approach without \acp{RNN} presented in this work (the Egocentric Map), demonstrates that the indiscriminate application of \acp{RNN}--as commonly employed in other works to address partial observability--can be suboptimal.
As explained in \Cref{sec:backg}, \acp{RNN} will implicitly construct belief states from the \ac{LiDAR} data, functionally similar to the explicit map representation in \textit{ego\_map}.
However, the explicit map offers superior interpretability and enables straightforward inter-agent communication through map merging.
By contrast, the latent space of \acp{RNN} lacks inherent interpretability, and merging these representations between agents presents significant challenges.




\section{Conclusions}
\label{sec:concl}

This work presents a framework for multi-agent communication-aware exploration of unknown indoor environments. 
A decentralized and scalable multi-agent system is considered within a constrained communication network to approximate real mission conditions.

The capabilities of the framework are demonstrated through the application of \ac{MARL} to train a group of \acp{UAV} to collaboratively explore an unknown \ac{GNSS}-denied indoor environment using high-fidelity game-engine simulation. Each \ac{UAV} agent is equipped with a \ac{LiDAR} sensor, which is used to build a local occupancy map. Each agent can also exchange data with neighbouring agents in a limited range. \ac{UAV} agents are trained in continuous action spaces to navigate under uncertainty using ND-POMDPs.

Through extensive ablation studies, this work systematically evaluates key factors, including training paradigms, reward functions, and network architectures. It also demonstrates that Curriculum-Learning significantly accelerates and stabilises training.
Furthermore, the resulting policies address critical limitations of prior work, such as reliance on discrete actions and centralized control, thereby laying a principled foundation for deploying learned strategies onto physical robotic systems.

Experimental results yield several key findings:
\begin{enumerate}[label=(\arabic*)]
\item A multi-agent perspective is essential.
      For instance, penalty-based collision avoidance resulted in agents learning not to move.
\item A \ac{SLAM}-RL integration proved to be more effective than standard \acp{RNN} for handling partial observability.
\item A specialized \ac{LiDAR} convolution network provided a 20\% performance gain, underscoring the value of domain-specific perception architectures.
\item A robust and accessible simulation framework was established using the Godot Engine, providing a lightweight, open-source platform with high-fidelity physics, validated with industry-grade tools on an \ac{HPC} cluster.
\item Training was successfully achieved via a structured Curriculum-Learning approach, enabling agents to master navigation in progressively complex environments.
\end{enumerate}

The work acknowledges inherent limitations, including the substantial sample complexity of \ac{MARL}, which required extensive training durations, and potential challenges in environment generalization.
Furthermore, the study focuses on a comparison of multi-agent learning paradigms rather than direct benchmarks against classical coverage algorithms (e.g. frontier-based).

Positioned as a stepping stone for future work, the research outlines a comprehensive trajectory for subsequent investigation.
Prospective directions include the following:

\begin{itemize}



\item \textbf{Improved sensors and observations}, including visual sensors and Visual-\ac{SLAM} support, dynamic rotation within the egocentric map for smoother \ac{UAV} trajectories~\cite{learning_cp_unkown_env_drl},
      and limiting the \ac{LiDAR} field-of-view from the unrealistic 360$^\circ$.

\item \textbf{Expansion to 3D} settings, adapting the formulations of agent kinematics, accounting for different information gains from observations and resulting beliefs, as well as the impact of communication limitations in mission performance.

\item \textbf{SLAM-RL}.
Experiment with different options, leveraging the commonalities, for instance Graph-\ac{SLAM} and \acp{GNN} on the RL side.








\item \textbf{Enhanced agent models and learning algorithms}, including improved credit assignment via value decomposition (e.g., \ac{QMIX}~\cite{qmix});
      hierarchical policies for multi-level decision-making; pre-training through imitation learning from expert demonstrations;
      competitive dynamics via game-theoretic self-play~\cite{self-play-survey}; and integrated planning with AlphaZero-inspired search algorithms such as Monte Carlo Tree Search.

\item \textbf{Temporal reasoning}.
The performance improvement observed when applying \acp{RNN} to \textit{ego\_map} likely stems from temporal reasoning about agent movement patterns, analogous to frame stacking in RL.
This could be done by applying 3D \acp{CNN}.

Given these considerations, the formulation with \ac{LiDAR} Convolutions may be preferable
due to its applicability to \ac{GNSS}-denied environments, architectural simplicity, and faster training
despite the 12.5\% reduction in final performance compared to the best \ac{RNN} configuration.
Future research on improved observation representations could further narrow this performance gap.

\end{itemize}

\section*{Acknowledgements}

The author wishes to thank HPCvLAB, whose computational resources made this work possible in its current form.
Providing access to the Deucalion~\cite{deucalion} and Cirrus~\cite{cirrus} \ac{HPC} Clusters.

This work is supported by the Portuguese Foundation for Science and Technology (FCT) under Grant 2023.04842.BD.

This work is also funded by national funds through FCT -- Fundação para a Ciência e a Tecnologia, I.P.,
under projects/supports UID/6486/2025 (\url{https://doi.org/10.54499/UID/06486/2025})
and UID/PRR/6486/2025 (\url{https://doi.org/10.54499/UID/PRR/06486/2025}).


\bibliographystyle{elsarticle-num}
\bibliography{main}


%

\end{document}